\documentclass{article}

\AtBeginDocument{%
  \addtolength\abovedisplayskip{-0.25\baselineskip}%
  \addtolength\belowdisplayskip{-0.25\baselineskip}%
  \addtolength\abovedisplayshortskip{-0.25\baselineskip}%
  \addtolength\belowdisplayshortskip{-0.25\baselineskip}%
}

     \usepackage[final]{neurips_2019}

\usepackage[utf8]{inputenc} 
\usepackage[T1]{fontenc}    
\usepackage{hyperref}       
\usepackage{url}            
\usepackage{booktabs}       
\usepackage{amsfonts}       
\usepackage{nicefrac}       
\usepackage{microtype}      
\usepackage{smile}
\title{Meta Learning with Relational Information \\ for Short Sequences}

\author{%
  Yujia Xie\\
  Georgia Tech\\
  \texttt{Xie.Yujia000@gmail.com} \\
\And
  Haoming Jiang\\
  Georgia Tech\\
  \texttt{jianghm@gatech.edu} \\
  \And
  Feng Liu\\
  Florida Atlantic University \\
  \texttt{FLIU2016@fau.edu}
  \And
  Tuo Zhao\\
  Georgia Tech \\
  \texttt{tuo.zhao@isye.gatech.edu}
  \And
  Hongyuan Zha\\
  Georgia Tech \\ 
  \texttt{zha@cc.gatech.edu}
}

\begin{document}

\maketitle


\begin{abstract}


This paper proposes a new meta-learning method -- named HARMLESS (HAwkes Relational Meta LEarning method for Short Sequences) for learning heterogeneous point process models from 
short event sequence data along with a relational network. Specifically, we 
propose a hierarchical Bayesian mixture Hawkes process model, which naturally incorporates the relational information among sequences into point process modeling. Compared with existing methods, our model can capture the underlying mixed-community patterns of the relational network, which simultaneously encourages knowledge sharing among sequences and facilitates adaptive learning for each individual sequence. We further propose an efficient stochastic variational meta expectation maximization algorithm that can scale to large problems. Numerical experiments on both synthetic and real data show that HARMLESS outperforms existing methods in terms of predicting the future events.


\end{abstract}
\vspace{-0.25in}
\section{Introduction}
\label{sec:intro}
\vspace{-0.1in}

Event sequence data naturally arises in analyzing the temporal behavior of real world subjects \citep{cleeremans1991learning}. These sequences often contain rich information, which can predict the future evolution of the subjects. 
For example, the timestamps of tweets of a twitter user reflect his activeness and certain state of mind, and can be used to show when he will tweet next time \citep{kobayashi2016tideh}. The job hopping history of a person usually suggests when he will hop next time \citep{xu2017learning}. 
Unlike usual sequential data such as text data, event sequences are always asynchronous and tend to be noisy \citep{ross1996stochastic}. Therefore specialized algorithms are needed to learn from such data. 

In this paper, we are interested in \textit{short} sequences, a type of sequence data that commonly appears in many real-world applications. Such data is usually short for two possible reasons. One is that the event sequences are short in nature, such as the job hopping history. Another is the observation window is narrow. For example, we are interested in the criminal incidents of an area after a specific regulation is published. 
Moreover, this kind of data usually appears as a collection of sequences, such as the timestamps of many user's tweets. 
Our goal is to extract information that can predict the occurrence of future events from a large collection of such short sequences.

Many existing literature considers medium-length or long sequences. They first model a sequence as a parametric point process, e.g., Poisson process, Hawkes process or their neural variants, and apply maximum likelihood estimation to find the optimal parameters \citep{ogata1999seismicity, rasmussen2013bayesian}. However, for short sequences, their lengths are insufficient for reliable inference.
One remedy is that we treat 
the collection of 
short sequences as independent identically distributed realizations of the same point process, since many subjects, e.g., Twitter users, often share similar behaviors. This makes the inference manageable. 
However, the learned pattern can be highly biased against certain individuals, especially the non-mainstream users, since this method ignores the heterogeneity within the collection.

An alternative  is to recast the problem as a multitask learning problem \citep{zhang2017survey} -- we target at multi-sequence analysis for multi-subjects. For each sequence, we consider a point process model that slightly deviates from a common point process model, i.e.,
$\tilde{f}_j = f_0 + f_j$,
where $f_0$ is the common model that captures the main effect, $\tilde{f}_j$ is the model for the $j$-th sequence, and $f_j$ is the relatively small deviation.
Such an assumption that there exists a universal common model cross all subjects, however, is still strong, since the subjects' patterns can differ dramatically. For example, the job hopping history of a software engineer and a human resource manager should have distinct characteristics. Furthermore, such method ignores the relationship of the subjects that usually can be revealed by side information. For example, a social network often shows community pattern \citep{girvan2002community} -- across the communities the variation of the subjects is large, while within the communities the variation is small. The connections in the social network, such as "follow" or retweet relationship in Twitter data, can provide us valuable information to identify such community pattern, but the aforementioned methods do not take into account such understanding to help analyzing subjects' behavior. 

To this end, we propose a HAwkes Relational Meta LEarning method for Short Sequence (HARMLESS), which can adaptively learn from a collection of  short sequence. 
More specifically, in a social network, each user often has multiple identities \citep{airoldi2008mixed}. For example, a Twitter user can be both a military fan and a tech fan. Both his tweet history and social connections are based on his identities. 
Motivated by above facts, we model each sequence as a hierarchical Bayesian mixture of Hawkes processes -- the weights of each Hawkes process are determined jointly by the hidden pattern of sequences and the relational information, e.g., social graphs. 

We then propose a variational meta expectation maximization algorithm to efficiently perform inference. Different from existing fully bayesian inference methods \citep{box2011bayesian, rasmussen2013bayesian, xu2017dirichlet}, we make no assumption on the prior distribution of the parameters of Hawkes process. Instead, when inferring for the Hawkes process parameters of the same identity for all the subjects, we perform a model-agnostic adaptation from a common model for this identity (\citet{finn2017model}, see section \ref{sec:model} for more details). 
This  is more flexible since it does not restrict to a specific form. 
We apply HARMLESS to both synthetic and real short event sequences, and achieve competitive performance. 





\noindent {\bf Notations}: 
Throughout the paper, the unbold letters denote vectors or scalars, while the bold letters denote the corresponding matrices or sequences. We refer the $k$-th entry of vector $a_i$ as $a_{i,k}$. We refer the $i$-th subject as subject $i$.




\vspace{-0.15in}
\section{Preliminaries}
\vspace{-0.1in}

We briefly introduce Hawkes Process and Model-Agnostic Meta Learning.

\textbf{Hawkes processes} \citep{hawkes1971spectra} is a doubly stochastic temporal point process $\mathcal{H}(\theta)$ with conditional intensity function $\lambda=\lambda(t; \theta, \bm{\tau})$ defined as
\vspace{-0.8pt}
\begin{align*}
\lambda(t; \theta, \bm{\tau}) = \mu + \sum_{\tau^{(j)}<t} g(t-\tau^{(j)}; \xi),
\end{align*}
where $\theta = \{\mu, \xi\}$, $g$ is the nonnegative impact function with parameter $\xi$, $\mu$ is the base intensity, and $\bm{\tau} = \{\tau^{(1)},\tau^{(2)},\cdots, \tau^{(M)}\}$ are the timestamps of the events occurring in a time interval $[0,t_{\rm{end}}]$. Function $g$ indicates how past events affect current intensity.  Existing works usually use pre-specified impact functions in parametric form, e.g., the exponential function in \citet{rasmussen2013bayesian, zhou2013learning} and the power-law function in \citet{zhao2015seismic}.

Hawkes process captures an important property of real-world events -- self-exciting, i.e., the past events always increase the chance of arrivals of new events. For example, selling a significant quantity of a stock can precipitate a trading flurry.
As a result, Hawkes process has been widely used in many areas, e.g., behavior analysis \citep{yang2013mixture,luo2015multi}, financial analysis \citep{bacry2012non}, and social network analysis \citep{blundell2012modelling,zhou2013learning}. 



\noindent \textbf{Model-Agnostic Meta Learning} (MAML, \citealp{finn2017model})
considers a set of tasks $\Gamma = \{\mathcal{T}_1, \mathcal{T}_2, \cdots, \mathcal{T}_N\}$, where each of the tasks only contains a very small amount of data which is not enough to train a model. We want to exploit the shared structure of the tasks, to obtain models that can perform well on each of the tasks. 
Specifically, MAML seeks to train a common model for all tasks. From optimization perspective, MAML solves the following problem,
\begin{align}
\min_{\theta} \sum_{\mathcal{T}_i\in \Gamma} \mathcal{F}_{\mathcal{T}_i} (\tilde{\theta}_i) \triangleq \min_{\theta} \sum_{\mathcal{T}_i\in \Gamma} \mathcal{F}_{\mathcal{T}_i} (\theta-\eta \mathcal{D}( \mathcal{F}_{\mathcal{T}_i}, \theta)), \label{eq:maml}
\end{align}
where $\mathcal{D}( \cdot, \cdot)$ is an operator, $\mathcal{F}_{\mathcal{T}_i}$ is the loss function of task $\mathcal{T}_i$, $\theta$ is the parameter of the common model,  and $\eta$ is the step size. 
Here, $\mathcal{D}( \mathcal{F}_{\mathcal{T}_i}, \theta)$ represents one or a small number of gradient update of $\theta$. For example, in cases of one gradient step, we take $\mathcal{D}( \mathcal{F}_{\mathcal{T}_i}, \theta) = \nabla_{\theta} \mathcal{F}_{\mathcal{T}_i}(\theta)$.
This optimization problem aims to find the common model that is expected to produce maximally effective behavior on that task after performing update $\theta-\eta\mathcal{D}( \mathcal{F}_{\mathcal{T}_i}, \theta)$.


Solving \eqref{eq:maml} using gradient descent involves computing the Hessian matrices, which is computationally prohibitive. To alleviate the computational burden, First Order MAML (FOMAML) \citep{finn2017model} and Reptile \citep{nichol2018first} are then proposed. FOMAML  drops the second order term in the gradient of \eqref{eq:maml}. Reptile further simplifies the computation by relaxing the original update with Hessian as a multi-step stochastic gradient descent updates. All three algorithms can be written in the form of \eqref{eq:maml} with operator $\mathcal{D}$ defined differently for different methods. Due to space limit, we defer the definition of $\mathcal{D}$ to Appendix \ref{sec:defD}.



\vspace{-0.15in}
\section{HAwkes Relational Meta LEarning for Short Sequences (HARMLESS)}
\label{sec:model}
\vspace{-0.1in}

\begin{wrapfigure}{R}{0.33\textwidth}
\vspace{-15pt}
  \begin{center}
    \includegraphics[width=0.22\textwidth]{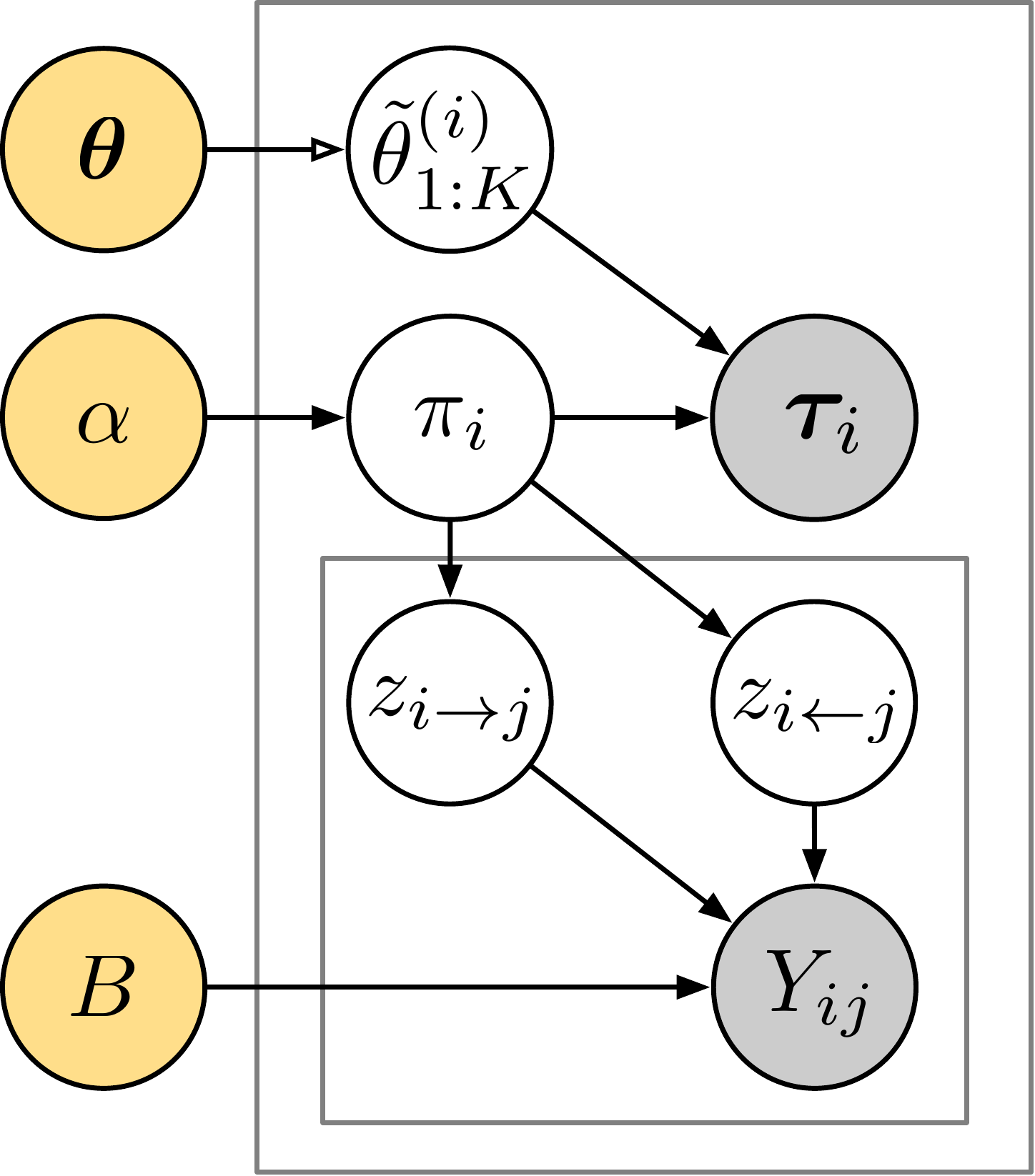}
  \end{center}
  \caption{\label{fig:mmb+seq}  Probabilisitic graph of the suggested model. The yellow nodes are parameters, white nodes are latent variables, and the gray nodes are observed variables. The solid arrows represent probabilistic mapping, while the hollow arrows represent the deterministic mapping. }
\vspace{-15pt}
\end{wrapfigure}

We next introduce the meta learning method for analyzing short sequences. Suppose we are given a collection of 
sequences $\bm{T} = \{\bm{\tau}_1, \bm{\tau}_2\, \cdots, \bm{\tau}_N\}$. We also know some extra relational information about the subjects. For example, in social networks, we can have information on who is friend of whom; in criminal data, we have the locations of the crimes, and crimes happen near each other often have Granger causality. Such relational information can be described as a graph $\mathcal{G} = (\mathcal{E},\mathcal{V})$, where $\mathcal{E}$ is the node set, $\mathcal{V}$ is the edge set. Denote its adjacency matrix as $\bm{Y}$.

Such social graphs often exhibit community patterns \citep{girvan2002community, xie2013overlapping}. Within the communities the variation of subjects are small, while across the communities the variation is large. Moreover, the communities are overlapping with each other, i.e., each subject may belong to multiple communities and  thus have multiple identities. 
The behaviors of the subject is based on the identities. 
Motivated by this observation, we first assign each subject a sum-to-one identity proportion vector $\pi_i\in [0,1]^K$, whose $k$-th entry represents the probability of subject $i$ having the $k$-th identity. In this way, we associate each subject with multiple identities rather than a single identity so that its different aspects is captured, which is more natural and flexible.

For the $k$-th identity of subject $i$, we adopt Hawkes process $\mathcal{H}(\tilde{\theta}_k^{(i)})$ to model the timestamps of the associated events. 
Denote the conditional intensity function of $\mathcal{H}(\tilde{\theta}_k^{(i)})$ as $\lambda(t;\tilde{\theta}_k^{(i)}, \bm{\tau}_i)$. For a Hawkes process $\mathcal{H}(\tilde{\theta}_k^{(i)})$, the likelihood \citep{laub2015hawkes} of a sequence $\bm{\tau}_i$ to appear in time interval $[0, t_{\rm{end}}]$ is 
\begin{align}
    \mathcal{L} (\tilde{\theta}_k^{(i)}; \bm{\tau}_i ) = \exp \Big(-\int_{0}^{t_{\rm{end}}} \lambda(t;\tilde{\theta}_k^{(i)}, \bm{\tau}_i) dt + \sum_{\tau_j<t_{\rm{end}}} \log \lambda(\tau_j; \tilde{\theta}_k^{(i)},\bm{\tau}_i)\Big). \label{eq:L_likelihood}
\end{align}
Here, the parameter $\tilde{\theta}_k^{(i)}$ is adapted from a common model with parameter $\theta_k$ using a relatively small model-agnostic adaptation, which we will elaborate in next section. 

The identity of the $i$-th subject is then a combination of the $K$ identities with identity proportion $\pi_i$, and the models for individual sequences are essentially mixtures of Hawkes process models. Denote $\mathcal{L}_i(\tilde{\theta}_k^{(i)}) = \mathcal{L}(\tilde{\theta}_k^{(i)}; \bm{\tau}_i)$. The likelihood for the $i$-th sequence $\bm{\tau}_i$ is
\begin{align}
p(\bm{\tau}_i) = \sum_{k=1}^K \pi_{i,k} \mathcal{L}_i(\tilde{\theta}^{(i)}_{k})  . \label{eq:pdf_seq}
\end{align}

Moreover, the connections of the subjects are also based on their identities. More specifically, for each connection to happen, one subject $i$ needs to approach another subject $j$, where the identities of subjects $i,j$ are based on $\pi_i, \pi_j$ respectively. Based on this observation, we adopt a Mixed Membership stochastic Blockmodel (MMB) \citep{airoldi2008mixed} to model the connections of the subjects. For each subjects pair $(i,j)$, denote the identity of subject $i$ when subject $i$ \textit{approaches} subject $j$ as random variable $z_{i \to j}$, and the identity of subject $j$ when $j$ \textit{is approached} by $i$ as $z_{i\leftarrow j}$. The probability of $z_{i\to j}$ represent the $k$-th identity is $\pi_{i,k}$, and the probability of $z_{i\leftarrow j}$ represent the $k$-th identity is $\pi_{j,k}$. 
 The probability of whether subject $i$ and $j$ have a connection is then a function dependent on this two identities - the random variable representing the existence of connection $Y_{ij}$ follows Bernoulli distribution with parameter $z_{i\to j}^T \bm{B} z_{i\leftarrow j}$, where $\bm{B}$ is a learnable parameter. 


\textbf{Generative process}: The above model can be summarized as the following generative process. 


\begin{itemize}
\item For each node $i$, 
\begin{itemize}
\item Draw a $K$ dimensional identity proportion vector $\pi_i\sim \text{Dirichlet}(\alpha)$.
\item Sample the $i$-th sequence $\bm{\tau}_i$ from the mixture of Hawkes processes described in \eqref{eq:pdf_seq}. 
\end{itemize}
\item For each pair of nodes $i$ and $j$,
\begin{itemize}
\item Draw identity indicator for the initiator $z_{i\to j} \sim \text{Categorical}(\pi_i)$
\item Draw identity indicator for the receiver $z_{i\leftarrow j} \sim \text{Categorical}(\pi_j)$
\item Sample whether there is an edge between $i$ and $j$, $Y_{ij} \sim \text{Bernoulli}(z_{i\to j}^T\bm{B} z_{i\leftarrow j})$.
\end{itemize}
\end{itemize}

Here, the observed variables are $\bm{\tau}_i$ and $Y_{ij}$. The parameters are  $\alpha$, $\tilde{\theta}_k^{(i)}$, and $\bm{B}$. The latent variables are $\pi_i$, $z_i$, $z_{i\to j}$ and $z_{i\leftarrow j}$. The graph model is shown in Figure \ref{fig:mmb+seq}.

\vspace{-0.15in}
\section{Variational Meta Expectation Maximization}
\vspace{-0.1in}

We now introduce our variational meta expectation maximization algorithm. This algorithm incorporates model-agnostic adaptation into variational expectation maximization. In the rest of the paper, we denote $\bm{z}_{\to}={\{z_{i\to j}\}_{i,j=1}^{N}}$, $\bm{z}_{\leftarrow}={\{z_{i\leftarrow j}\}_{i,j=1}^{N}}$, $\bm{\tilde{\theta}} = \{\tilde{\theta}_k^{(i)}\}_{i=1,k=1}^{N,K}$.


To ease the computation we add one more latent variable $\bm{z}$. For the $i$-th sequence, we sample $z_i \sim \text{Categorical}(\pi_i)$. We regard $\bm{\tau}_i$ as a Hawkes process with parameter $\theta_{z_i}^{(i)}$. Note that this is equivalent to the mixture of Hawkes process described in previous section, since $p(\bm{\tau}_i) = \sum_k p(z_i=k) \mathcal{L}_i(\tilde{\theta}_{z_i}^{(i)})=\sum_k \pi_{i,k} \mathcal{L}_i(\tilde{\theta}_{k}^{(i)})$. This can ease the computation because now the update for $\bm{\pi}$ has close form.

\textbf{Variational E step.} The goal is to find an approximation of the following posterior distribution
\begin{align*}
p( \bm{z} ,\bm{z}_{\rightarrow},\bm{z}_{\leftarrow}, \bm{\pi} | \bm{T},\bm{Y},\alpha,\bm{\tilde{\theta}} , \bm{B}).
\end{align*}

We aim to find a distribution $q(\bm{z} , \bm{z}_{\rightarrow},\bm{z}_{\leftarrow}, \bm{\pi})$ that minimizes the Kullback-Leibler (KL) divergence to the above posterior distribution. This can be achieved by maximizing the Evidence Lower BOund (ELBO, \citealp{blei2017variational}),
\begin{align}
    \max_{q \in Q} \mathbb{E}_q [\log p( \bm{z} ,\bm{z}_{\rightarrow},\bm{z}_{\leftarrow}, \bm{\pi}, \bm{T},\bm{Y})] - \mathbb{E}_q[\log q(\bm{z} , \bm{z}_{\rightarrow},\bm{z}_{\leftarrow}, \bm{\pi})],  \label{eq:elbo}
\end{align}
where $Q$ is a properly chosen distribution space.
We adopt $Q$ as the mean-field variational family, i.e.,
\begin{align*}
    q(\bm{z} , \bm{z}_{\rightarrow},\bm{z}_{\leftarrow}, \bm{\pi}) =  q_1(\bm{\pi})\prod_i q_2(z_i)\prod_j q_3(z_{i\to j}) q_4(z_{i\leftarrow j}).
\end{align*}
where $q_1(\pi_i)$ is the Probability Density Function (PDF) of $\text{Dirichlet}(\beta_i)$, $q_2(z_i)$ is the Probability Mass Function (PMF) of $\text{Categorical}(\gamma_i)$, $q_3(z_{i\to j})$ is the PMF of $\text{Categorical}(\phi_{ij})$, $q_4(z_{i\leftarrow j})$ is the PMF of $\text{Categorical}(\psi_{ij})$, and $\beta_i$, $\gamma_i$, $\phi_{ij}$, $\psi_{ij}$ are variational parameters. By some derivation (see Appendix \ref{sec:derEM} for detail), the updates for the variational parameters for solving problem \eqref{eq:elbo} are
    \begin{align}
    & \beta_{i,k} \leftarrow \alpha_k + \gamma_{i,k} + \sum_{j=1}^N \phi_{ij,k}+\sum_{j=1}^N \psi_{ij,k}, \label{eq:update_alpha1_comp} \\
    & \gamma_{i,k} \leftarrow  e^{\mathbb{E}_q [\log \pi_{i,k}]} \mathcal{L}_i(\tilde{\theta}_k^{(i)} ), \quad \gamma_{i,k} \leftarrow \frac{\gamma_{i,k}}{\sum_{\ell} \gamma_{i,\ell}}, \label{eq:update_gamma2_comp} \\
    & \phi_{ij,k} \leftarrow e^{\mathbb{E}_q [\log \pi_{i,k}]} \prod_{\ell=1}^K \left(B_{k\ell}^{Y_{ij}} (1-B_{k\ell})^{1-Y_{ij}}  \right)^{\psi_{ij,\ell}}, \quad \phi_{ij,k} \leftarrow \frac{\phi_{ij,k}}{\sum_{\ell} \phi_{ij,\ell}}, \label{eq:update_phi2_comp} \\
    & \psi_{ij,\ell} \leftarrow e^{\mathbb{E}_q [\log \pi_{j,\ell}]} \prod_{k=1}^K \left((B_{k\ell})^{Y_{ij}} (1-B_{k\ell})^{1-Y_{ij}}  \right)^{\phi_{ij,k}}, \quad \psi_{ij,\ell} \leftarrow \frac{\psi_{ij,\ell}}{\sum_{k} \psi_{ij,k}}, \label{eq:update_psi2_comp} 
    \end{align}
where  $\mathbb{E}_q [\log \pi_{i,k}] =  f_{\rm{dg}}(\beta_{i,k})-f_{\rm{dg}}(\sum_\ell \beta_{i,\ell})$, and $f_{\rm{dg}}(\cdot)$ is the digamma function. 
    
 \textbf{Meta inference for $\bm{\theta}$ and $\bm{\tilde{\theta}}$.}
 Recall that the Hawkes parameter of the $k$-th identity of subject $i$ is $\tilde{\theta}_{k}^{(i)}$.  Instead of specifying that $\tilde{\theta}_{k}^{(i)}$ is sampled from a prior distribution, we adapt the $k$-th common model $\mathcal{H}(\theta_{k})$ to sequence $i$ using MAML-type updates,
\begin{align}
\tilde{\theta}^{(i)}_{k} = \theta_{k} - \eta \mathcal{D} (\log\mathcal{L}_i ,\theta_{k}). \label{eq:maml_adapt}
\end{align}
Since MAML-type algorithms only perform one or few updates from the common model, the adapted individual models  with parameter $\tilde{\theta}^{(i)}_{k}$ within one community is close to each other, which meets our expectation that the within-community variation should be small.

The gradient descent step on the log-likelihood of $\bm{\theta}$ can then be written as 
    \begin{align}
    & \theta_k \leftarrow \theta_k + \eta_{\bm{\theta}} \nabla_{\theta_k} \left(\sum_{i=1}^N  \gamma_{i,k} \log \mathcal{L}_i(\theta_k -\eta \mathcal{D}(\log \mathcal{L}_i,\theta_k)) \right), \label{eq:update_theta_comp} 
    \end{align}
where $\eta_{\bm{\theta}}$ is the step size. In this algorithm, we only need to estimate the common models with parameter $\theta_k$, $k=1,2,\cdots, K$ instead of all individual models. After we obtain $\theta_k$, the individual models can be easily obtained from Equation \eqref{eq:maml_adapt}. 
 
 \textbf{M step.} We perform maximum likelihood estimation to $\alpha$ and $B$, The updates are as follows,
    \begin{align}
    & \alpha_{k} \leftarrow \alpha_{k} + 
    \eta_{\alpha} \left( N \big( f_{\rm{dg}}(\sum_\ell \alpha_{\ell}) - f_{\rm{dg}}(\alpha_{k})\big)+ \sum_{i=1}^N\big(  f_{\rm{dg}}(\beta_{i,k})-f_{\rm{dg}}(\sum_l \beta_{i,\ell})\big)\right), \label{eq:update_alpha_comp} \\
    & B_{k\ell} \leftarrow \frac{\sum_{ij} Y_{ij} \phi_{ij,k} \psi_{ij,\ell}}{\sum_{ij} \phi_{ij,k} \psi_{ij,\ell}}, \label{eq:update_B_comp}
    \end{align}
where $\eta_{\alpha}$ is the step size. The detailed derivation can be found in Appendix \ref{sec:derEM}. 

\textbf{Algorithm.} We perform updates \eqref{eq:update_alpha1_comp}-\eqref{eq:update_psi2_comp}, \eqref{eq:update_theta_comp}-\eqref{eq:update_B_comp} iteratively until convergence. 
Note that the updates can also be implemented in stochastic fashion -- at each iteration, we sample a mini-batch of sequences, and update their associated parameters  \citep{hoffman2013stochastic}.



\vspace{-0.15in}
\section{Experiments}
\vspace{-0.1in}

We first briefly introduce oue experiment settings. 

\textbf{Impact function.} Following \citet{rasmussen2013bayesian, zhou2013learning}, we choose exponential impact function $g(t; \{\delta, \omega\}) = \delta \omega e^{-\omega t}$. The conditional intensity function is
\begin{align}
\lambda(t; \theta,\bm{\tau}) = \lambda(t; \{\mu, \delta, \omega\},\bm{\tau}) = \mu + \sum_{\tau^{(m)}<t} \delta \omega e^{-\omega(t-\tau^{(m)})}, \label{eq:lambda}
\end{align}
where $\delta$ and $\omega$ are parameters.
Note that each Hawkes process model only contains three parameters, $\mu$, $\delta$, and $\omega$. This is because we target at short sequence. To avoid overfitting, each individual models cannot have too many parameters. 

\textbf{Regularized likelihood function.} Substitute Eq. \eqref{eq:lambda} into Eq. \eqref{eq:L_likelihood}, we have
\begin{align*}
    \mathcal{L} (\theta; \bm{\tau}) = \exp \Big( -\mu t_{\rm{end}} - \hspace{-0.125in}\sum_{\tau^{(n)} < t_{\rm end}}\hspace{-0.125in}\Big( \delta (1-e^{-\omega(t_{\rm{end}}-\tau^{(n)})})-\log \big(\mu + \hspace{-0.125in}\sum_{\tau^{(m)}<\tau^{(n)}}\hspace{-0.125in}\delta \omega e^{-\omega(\tau^{(n)}-\tau^{(m)})}\big) \Big)\Big).
\end{align*}
To keep the parameters non-negative, in practice we replace $\log \mathcal{L}_i (\tilde{\theta}_k^{(i)}) $ with a regularized log-likelihood in update \eqref{eq:update_theta_comp},
\begin{align}
\mathcal{Q}_i(\tilde{\theta}_k^{(i)}) \triangleq \log \mathcal{L}_i (\tilde{\theta}_k^{(i)}) + \nu \mathcal{R} (\tilde{\theta}_k^{(i)}) \triangleq \log \mathcal{L}_i (\tilde{\theta}_k^{(i)}) + \nu \big(  \log(\tilde{\mu}_k^{(i)}) + \log(\tilde{\alpha}_k^{(i)}) + \log(\tilde{\omega}_k^{(i)})\big),  \label{eq:obj}
\end{align}
where $\tilde{\theta}_k^{(i)} = \{\tilde{\mu}_k^{(i)},\tilde{\alpha}_k^{(i)},\tilde{\omega}_k^{(i)}\}$ is the parameter of the $i$-th Hawkes process of the $k$-th identity, $\nu$ is a regularization coefficient. 


\textbf{Evaluation metric.}  We hold out the last timestamp of each sequence, and split the hold-out timestamps into a validation set and a test set. 
Another option to do validation and test on event sequence data is to hold out the last two timestamps -- we first use the former ones to do validation, then train a new model together with the validation timestamps, and finally report the test result based on the later ones. However, this is not suitable here.
This is because the sequences we adopt for experiments are usually very short, sometimes even no more than 5 events in one sequence. As a result, the models trained without or with validation timestamps, e.g., using 3 or 4 timestamps, can be significantly different, which makes the validation procedure very unreliable. 

We report the Log-Likelihood (LL) of the test set. More specifically, for each sequence $\bm{\tau}_i = \{\tau_i^{(1)}, \tau_i^{(2)}, \cdots, \tau_i^{(M_i)}\}$ and parameter $\bm{\theta}$, the likelihood of next arrival $\tau_i^{(M_i+1)}$ is
 \begin{align*}
   \tilde{ \mathcal{L}}_{i}  = \sum_{k=1}^K \gamma_{i,k}  \lambda \big(\tau_i^{(M_i+1)}; \tilde{\theta}_{k}^{(i)} , \bm{\tau}_i)\big) \exp \Big( -\int_{\tau_i^{(M_i)}}^{\tau_i^{(M_i+1)}} \lambda(t ; \tilde{\theta}_{k}^{(i)} , \bm{\tau}_i) ~ dt\Big).
\end{align*}
The reported score is the averaged $\log \tilde{ \mathcal{L}}_{i}$ over  subjects. More details can be found in Appendix \ref{sec:derEv}.

To estimate of the variance of the estimated log-likelihood, we adopt a multi-split procedure for evaluation. First,  we train $m$ candidate models with different hyper-parameters. Then we repeat the following procedure for $30$ times: 1). Randomly split a validation set and a test set; 2). Pick a model with highest log-likelihood on the validation set from the $m$ candidate models; 3). Compute the log-likelihood on the test set. Accordingly, we obtain $30$ estimates of the log-likelihood. We then report the mean and standard error of the $30$ estimates. 

\textbf{Baselines.} We adopt four baselines as follows.

\noindent{$\diamond$ \it MLE-Sep}: We consider each sequence as a realization of an individual Hawkes process. We perform Maximum Likelihood Estimation (MLE) on each sequence separately, and obtain $N$ models for $N$ sequences. 

\noindent{$\diamond$ \it MLE-Com}: We consider all sequences as realizations of the same Hawkes process and learn a common model by MLE. 

\noindent{$\diamond$ \it DMHP} \citep{xu2017dirichlet}: We model sequences as a mixture of Hawkes processes with a Dirichlet distribution as the prior distribution of the mixtures.

\noindent{$\diamond$ \it MTL}: We perform multi-task learning as described in Section \ref{sec:intro}. More specifically, we adopt Hawkes process model for $f_0$ and $\tilde{f}_j$. Denote the parameters of $f_0$ and $\tilde{f}_i$ as $\rho_0=[\mu_0, \delta_0, \omega_0]^T$ and $\rho_i=[\mu_i, \delta_i, \omega_i]^T$, respectively. We solve 
\begin{align*}
\max_{\rho_0, \rho_i}\sum_{i=1}^N \left( \mathcal{Q}_i(\rho_i) + \nu_{\rm{mtl}} \norm{\rho_i-\rho_0}_2 \right),
\end{align*}
where $\norm{\rho_i-\rho_0}_2$ is the $\ell_2$ norm regularizer of $\rho_i-\rho_0$ to promote the difference between $f_0$ and $f_j$ to be small, $\nu_{\rm{mtl}}$ is a tuning parameter, and $\mathcal{Q}_i(\cdot)$ is the function defined in Eq. \eqref{eq:obj}. 


\textbf{Parameter Tuning.} The detailed tuning procedure and detailed settings of each experiment can be found in Appendix \ref{sec:set_exp}.




\begin{wraptable}{R}{9.3cm}
\vspace{-0.13in}
	\caption{\label{tab:graphs} Visualizations of identities by HARMLESS(MAML).}
	\vspace{-0.05in}
	\begin{tabular}{m{0.25cm} m{1.9cm}m{1.5cm}m{1.5cm}m{1.8cm}}
		\hline
		~$S$ & Ground Truth& ~~~~$K_{0}=3$&  ~~~~$K_{0}=6$& ~~~~$K_{0}=10$ \\
		\hline
		 $0.5$
		& \includegraphics[width=0.143\textwidth]{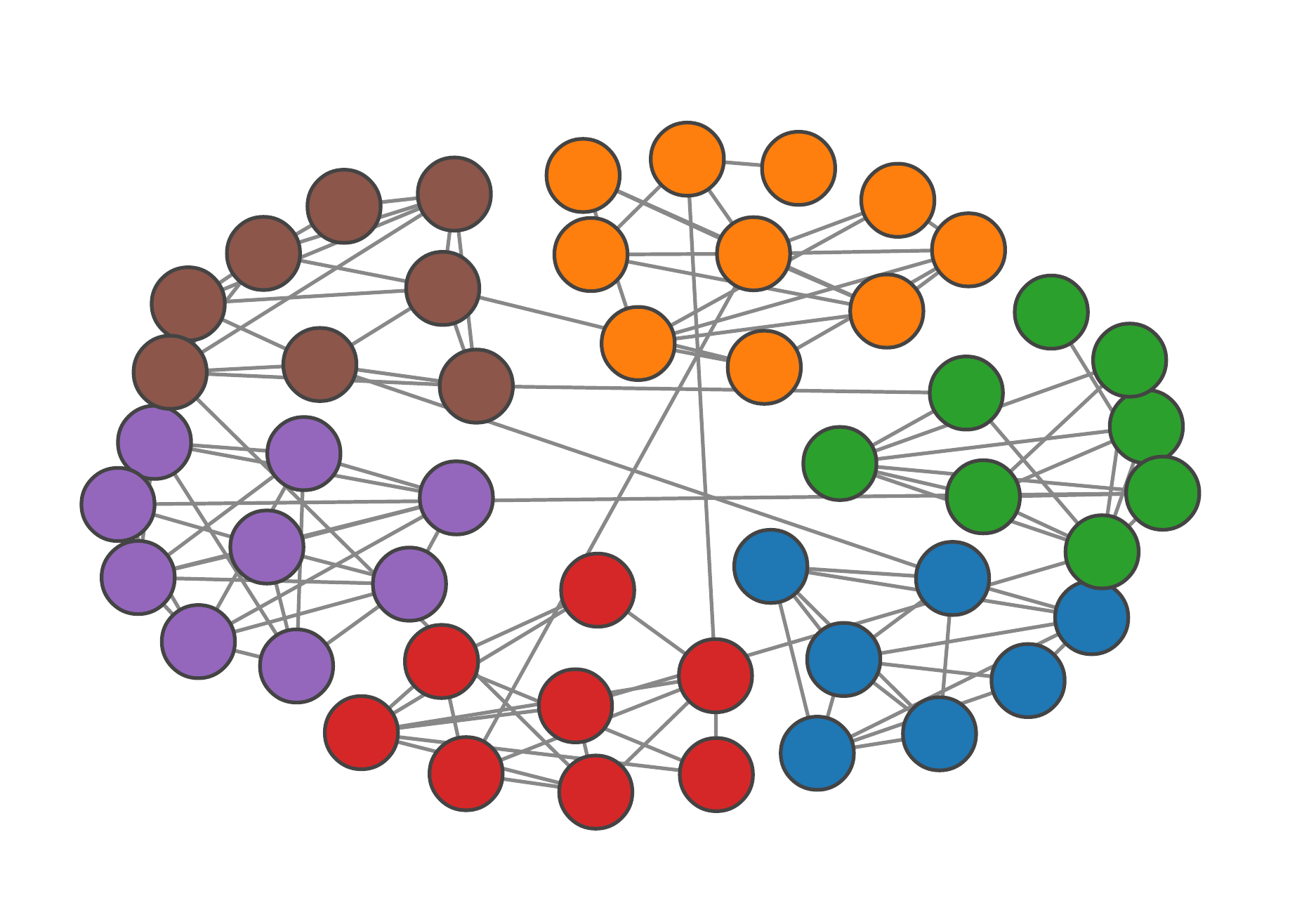} 
		&\includegraphics[width=0.143\textwidth]{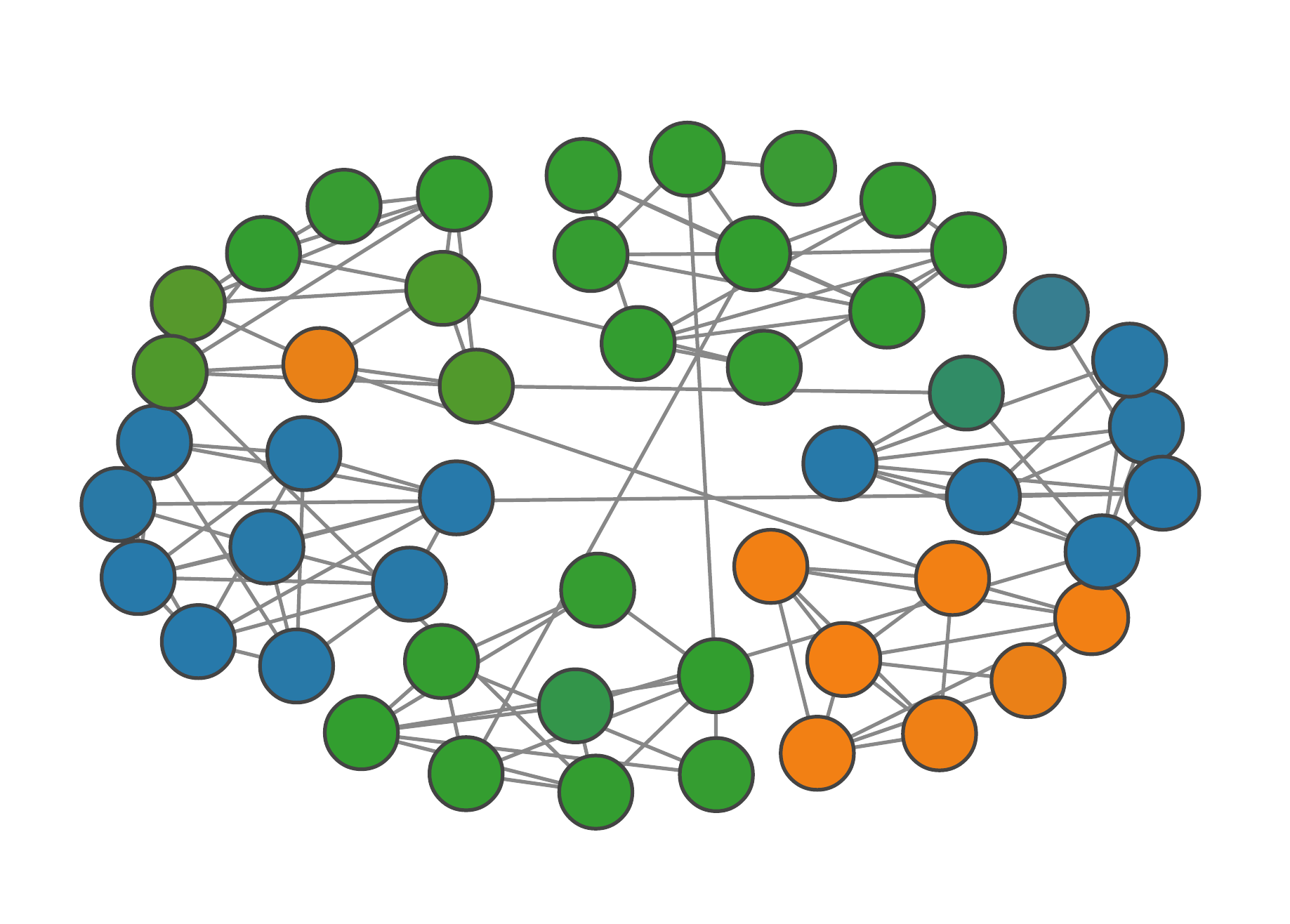}
		& \includegraphics[width=0.143\textwidth]{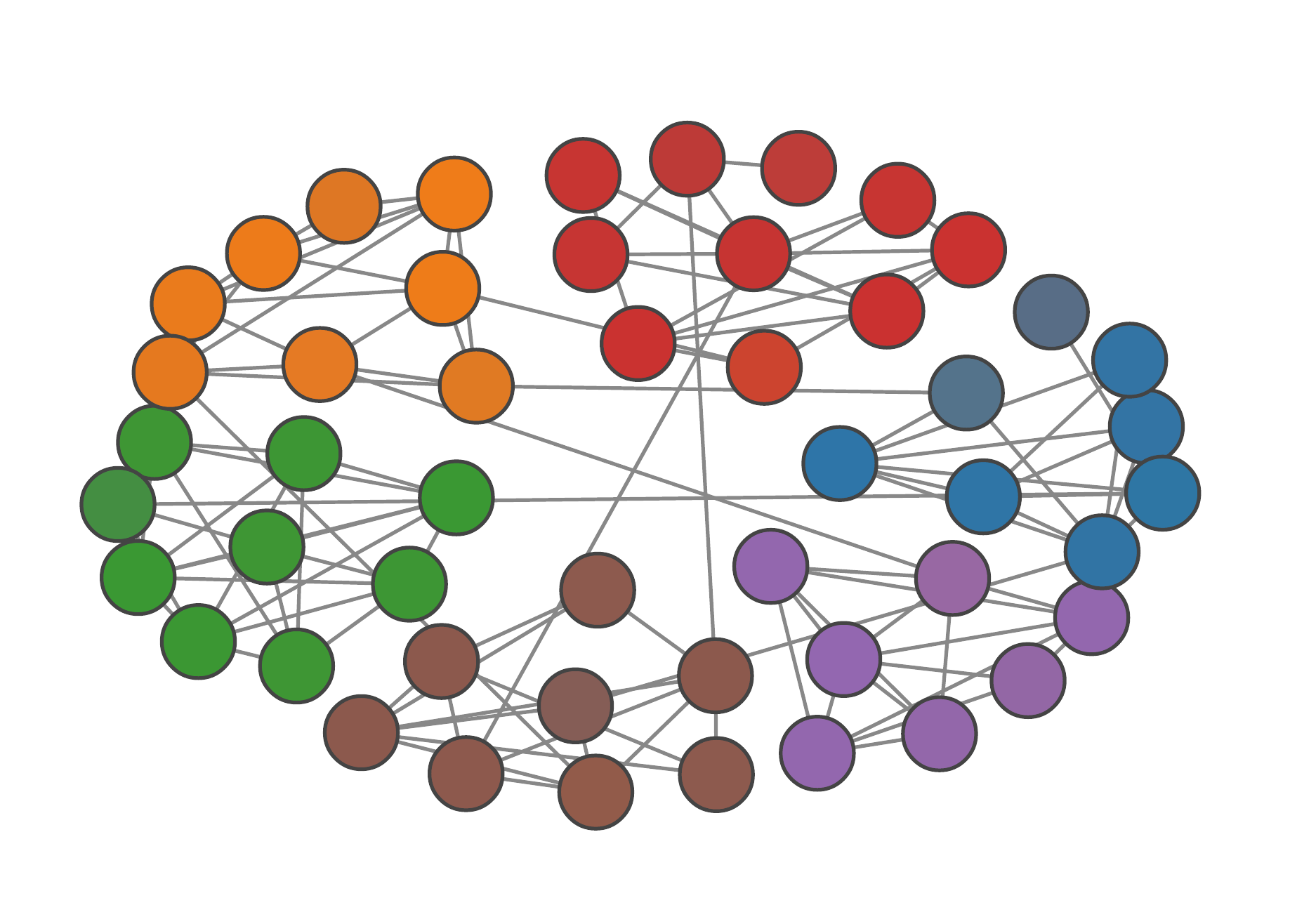}
		& \includegraphics[width=0.143\textwidth]{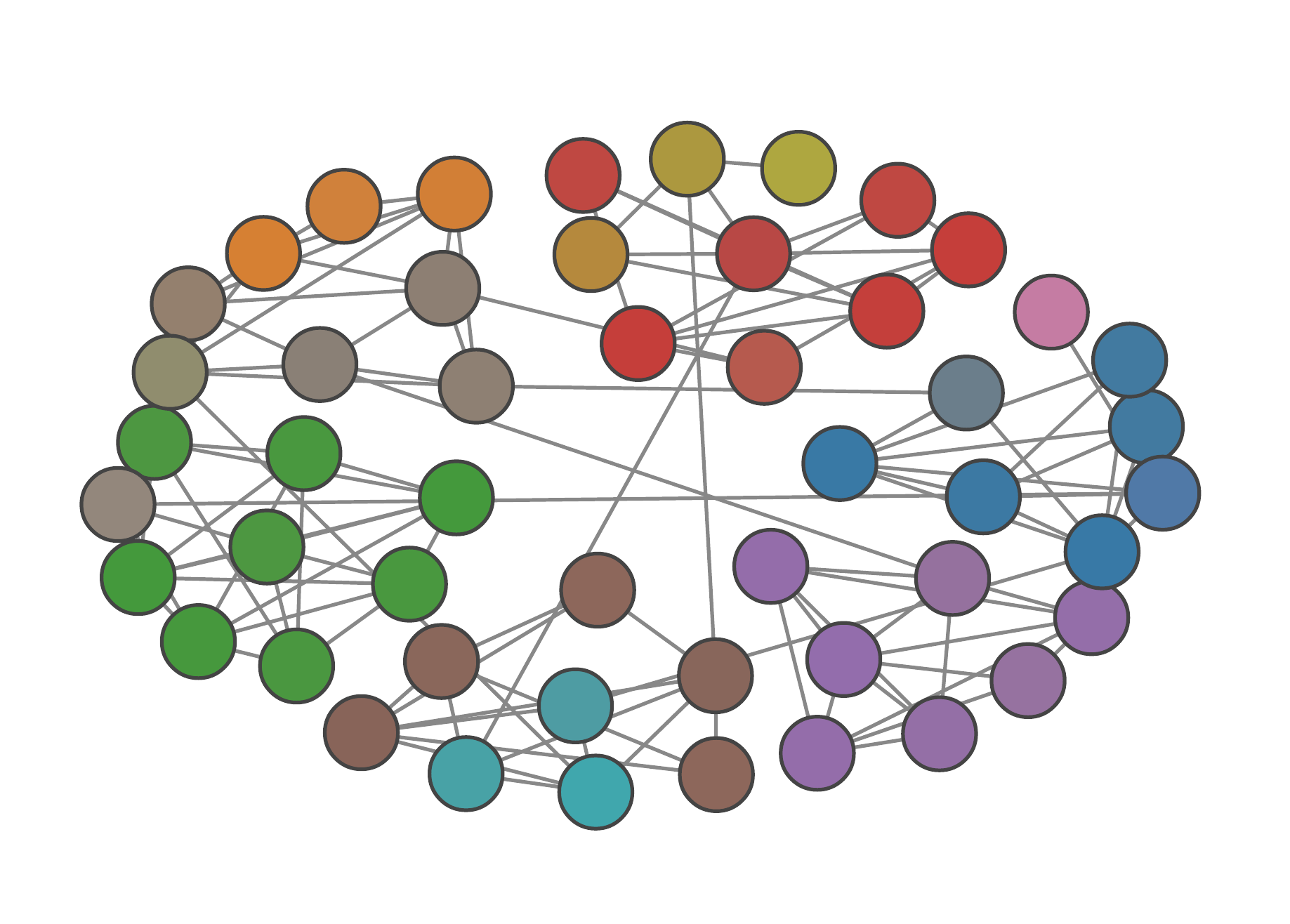} \\ 
		 $1.0$
		& \includegraphics[width=0.143\textwidth]{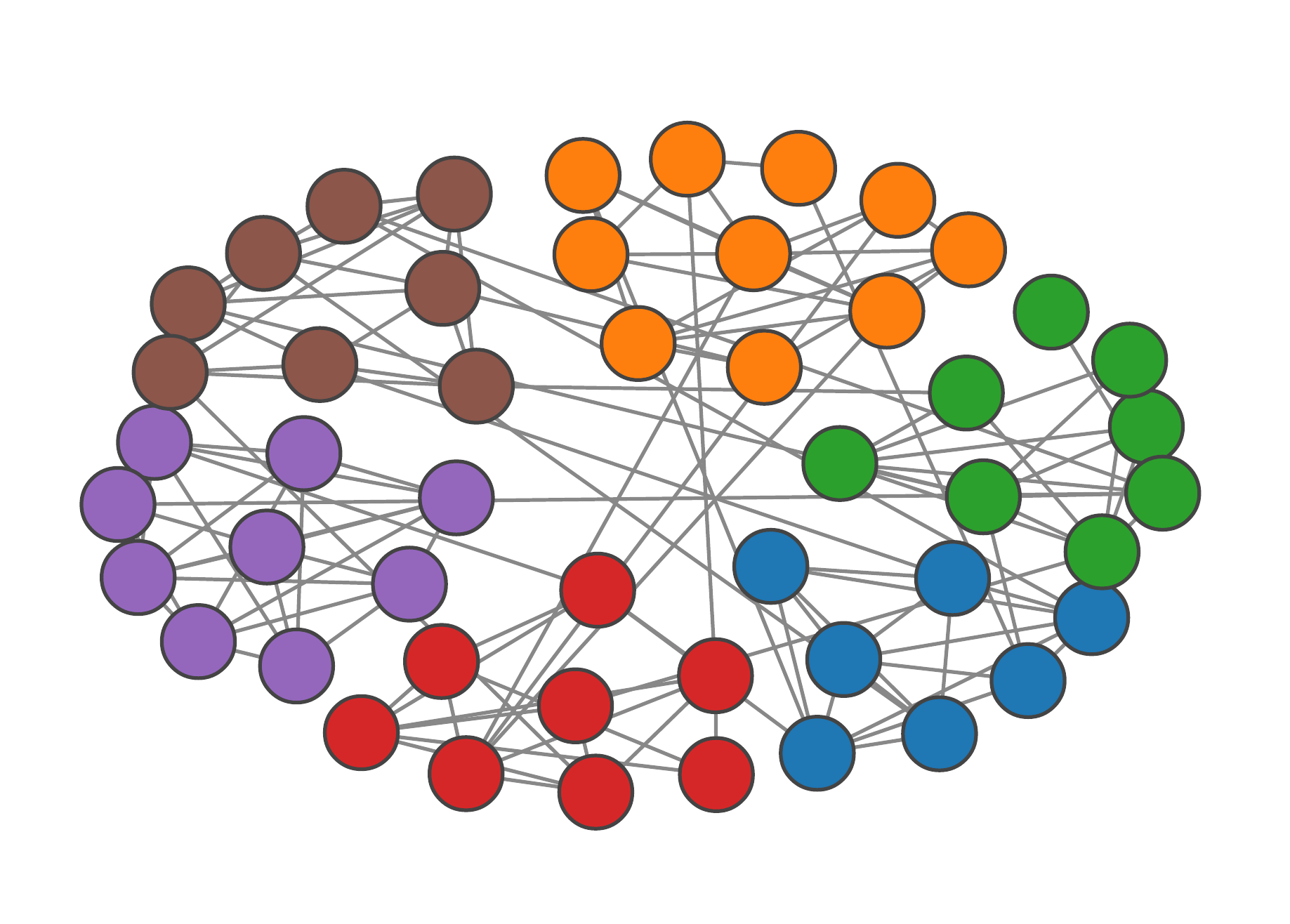} 
		&\includegraphics[width=0.143\textwidth]{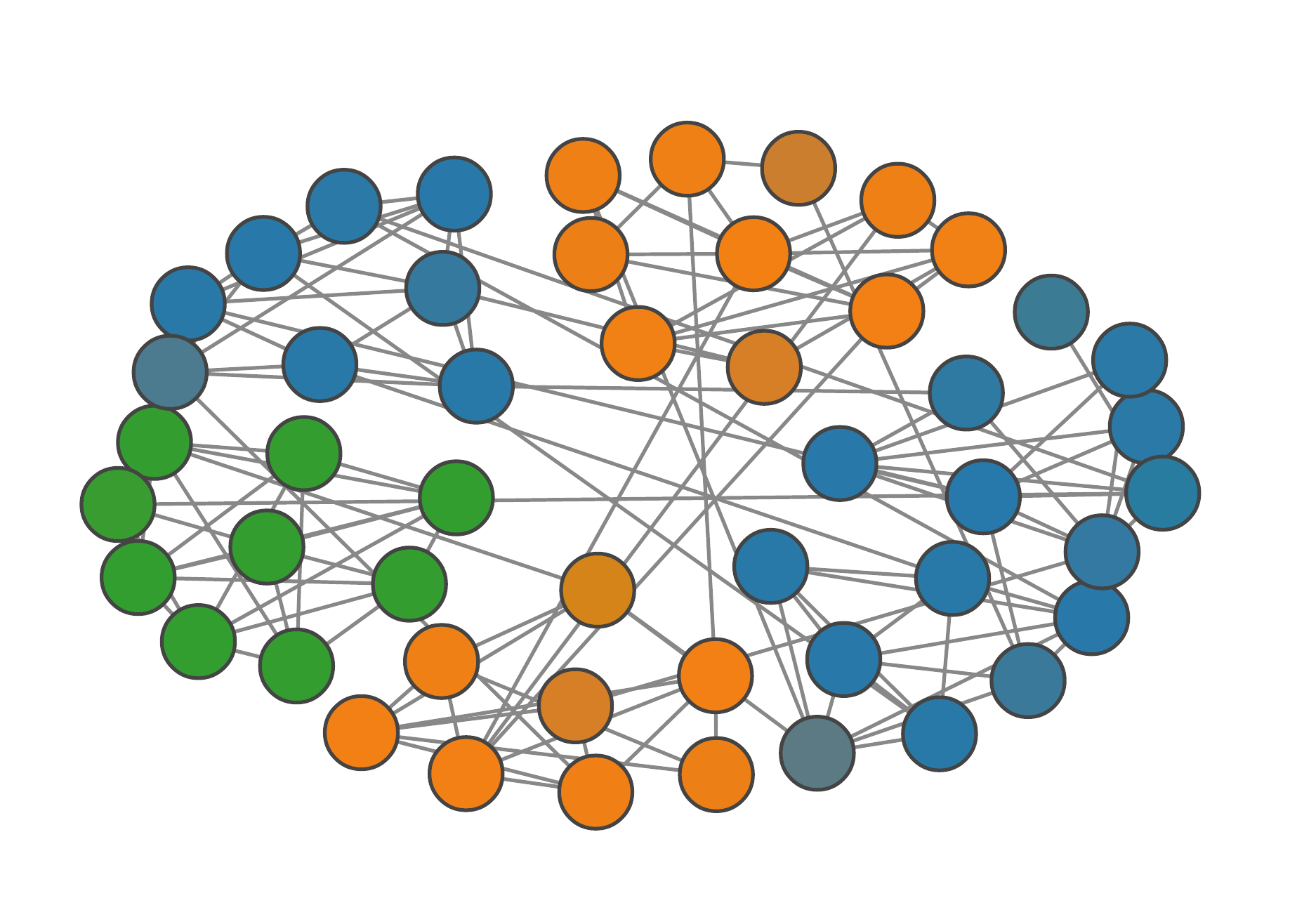}
		& \includegraphics[width=0.143\textwidth]{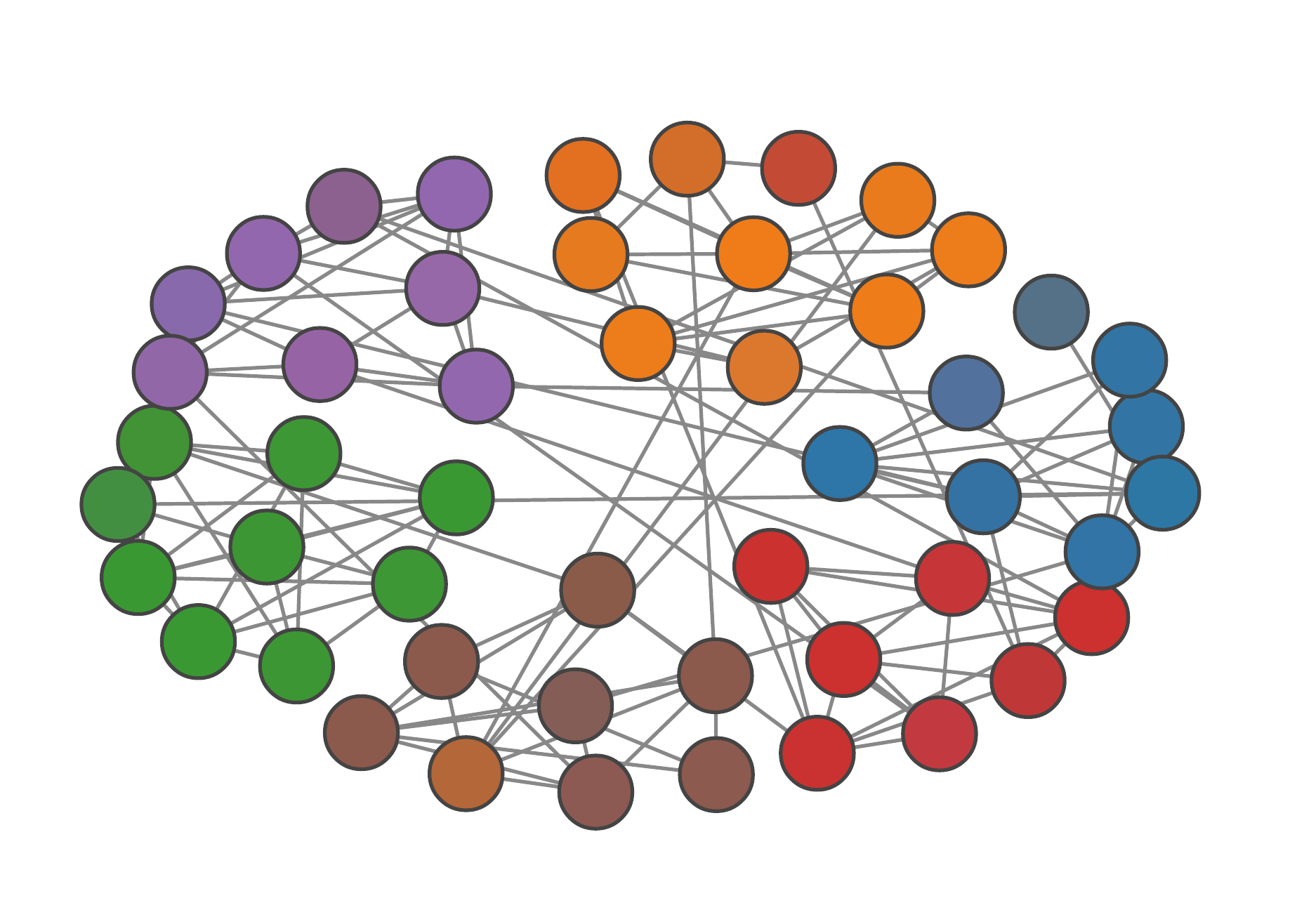}
		& \includegraphics[width=0.143\textwidth]{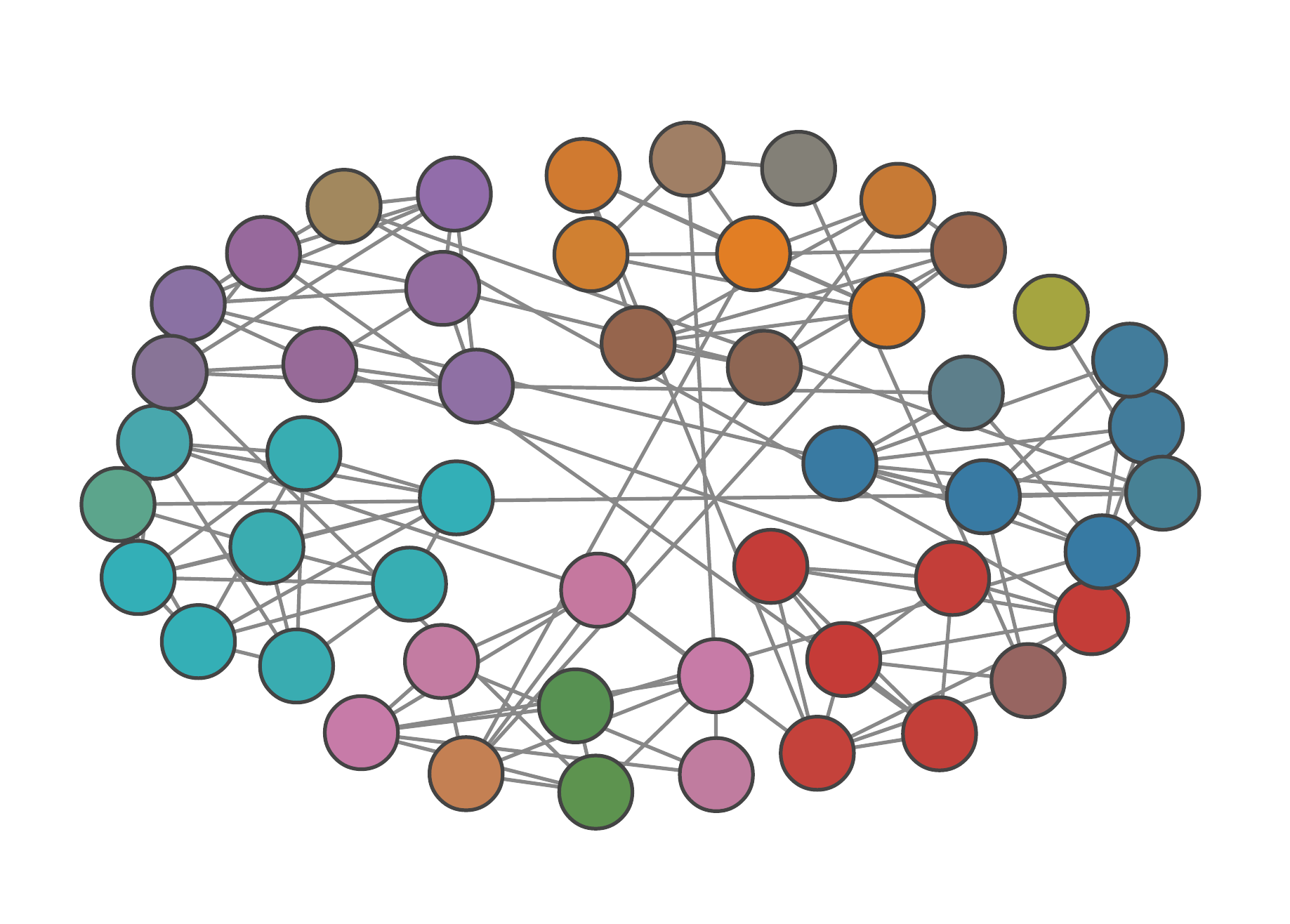} \\ 
		 $2.0$
		& \includegraphics[width=0.143\textwidth]{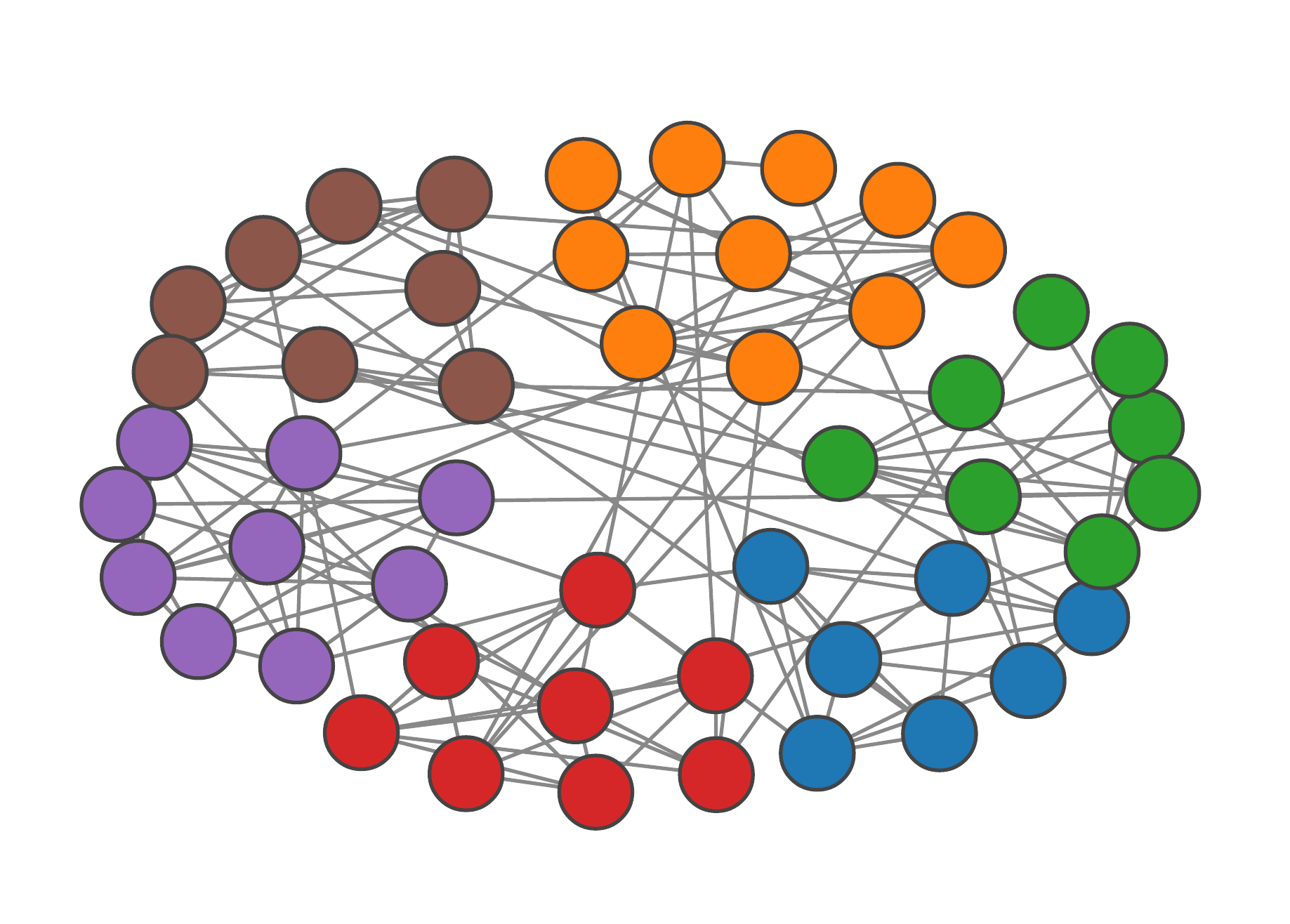} 
		&\includegraphics[width=0.143\textwidth]{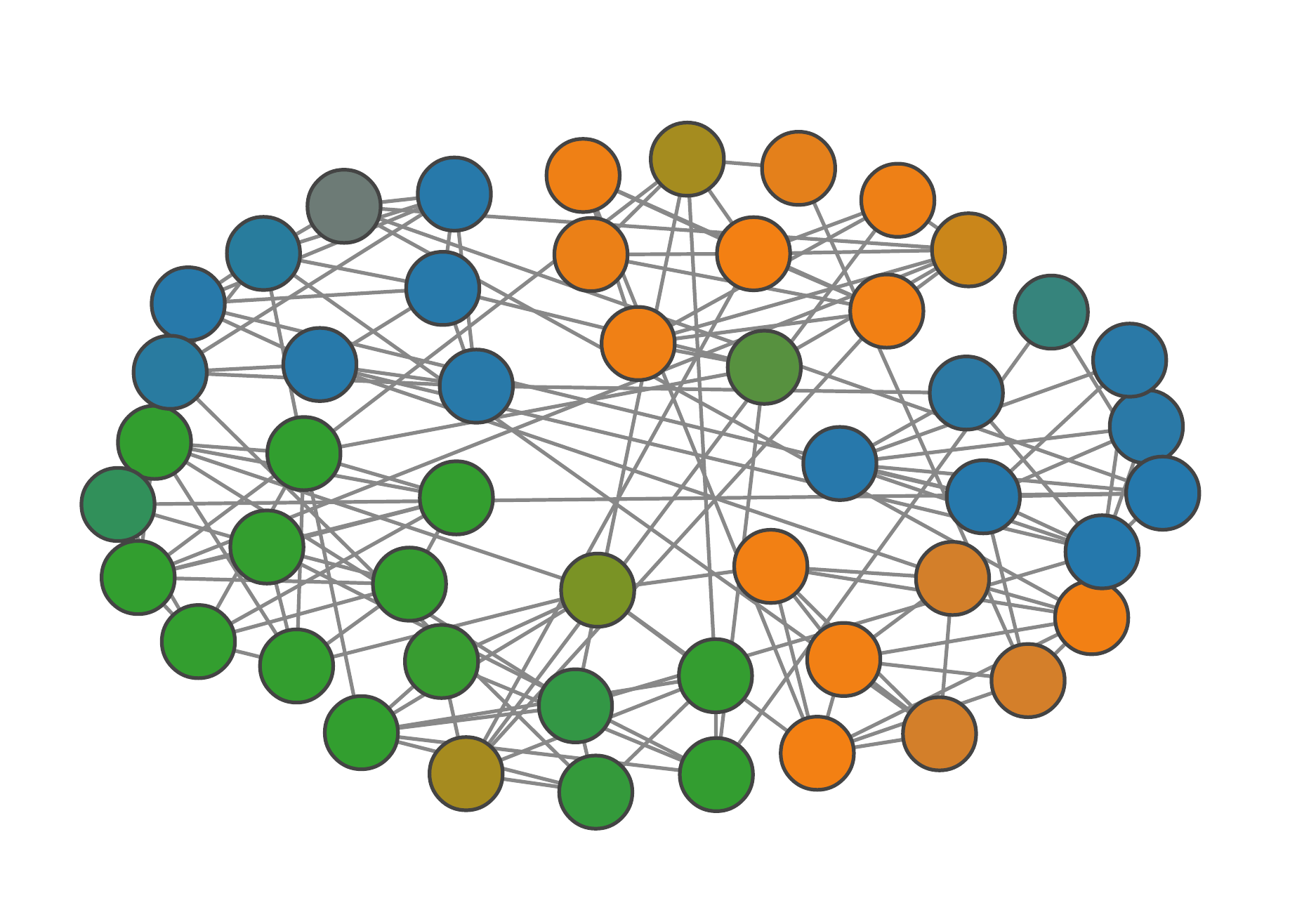}
		& \includegraphics[width=0.143\textwidth]{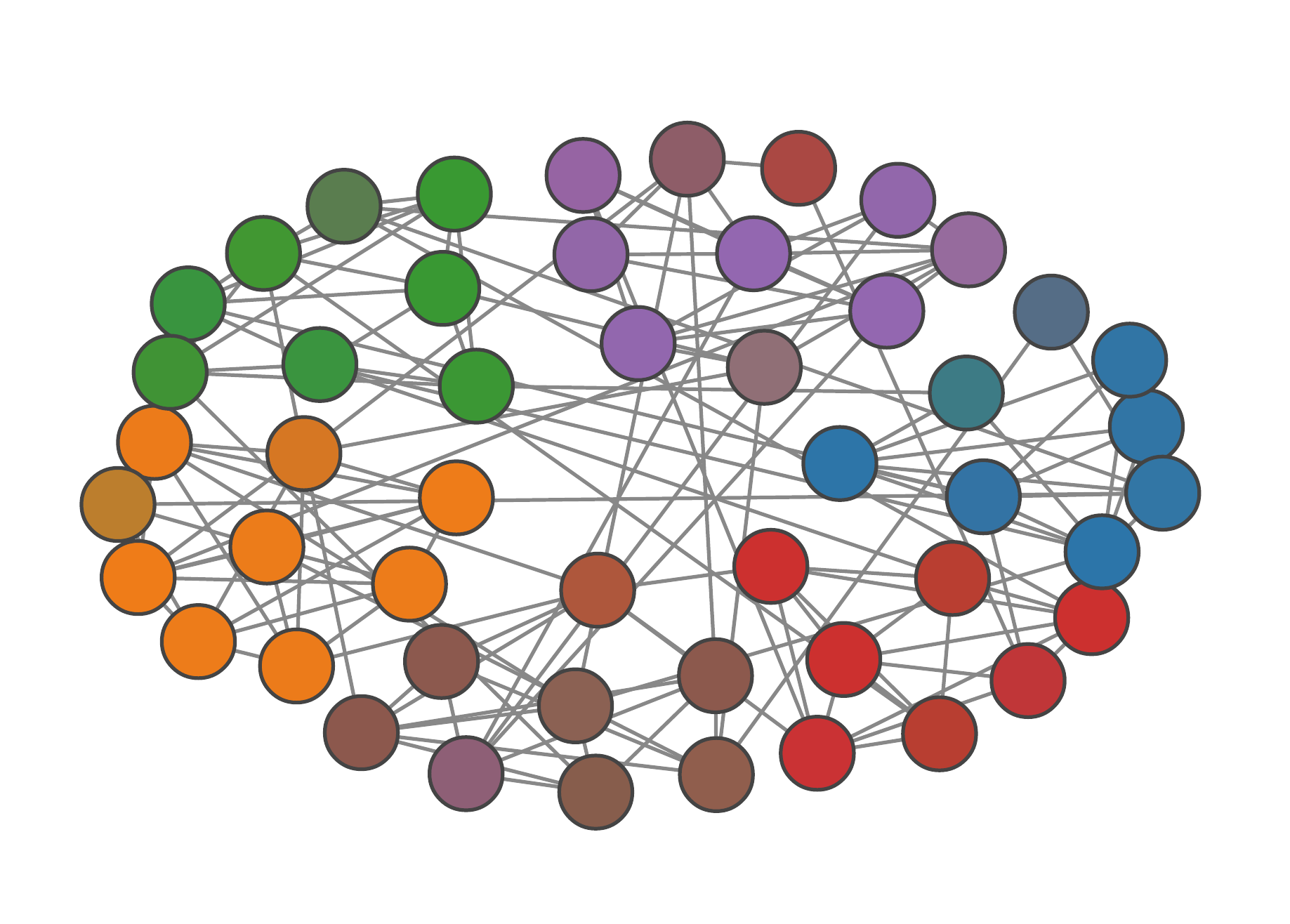}
		& \includegraphics[width=0.143\textwidth]{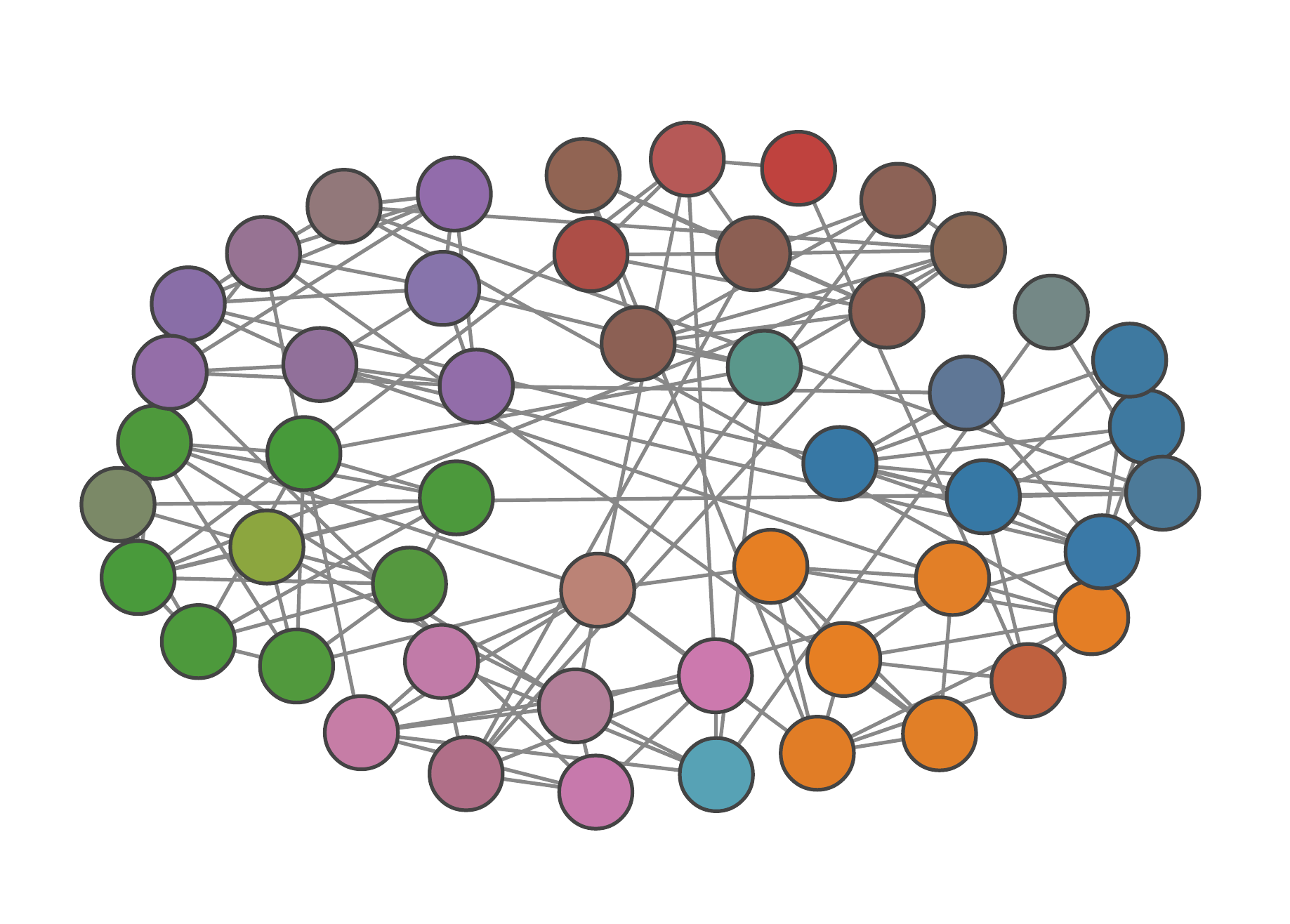} \\ 
		\hline
	\end{tabular}
\vspace{-15pt}
\end{wraptable} 

\vspace{-0.15in}
\subsection{Synthetic Data}
\vspace{-0.1in}

\textbf{Data generation.}
We generate a dataset of 50 nodes with $K=6$ communities. For each community, we generate Hawkes meta parameters $\theta_{k}=\{\mu_k,\delta_k,\omega_k\}$ using the following uniform distributions: 
\begin{align*}
	& \mu_k \sim \text{Uniform}(0.15,10), \quad 
	 \delta_k \sim \text{Uniform}(0.15,0.85), \quad 
	 \omega_k \sim \text{Uniform}(1,10)
\end{align*}
We set $\alpha=\mathbf{1}_K$, i.e., the entries of $\alpha$ is all one.
Then for the $i$-th node, the identity proportion $\pi_i$ is sampled from $\text{Dirichlet}(\alpha)$ and the membership indicator $z_i$ from the corresponding categorical distribution $ \text{Categorical}(\pi_i)$. 
Based on $z_i$, we then generate the Hawkes parameters $\tilde{\theta}_{z_i}^{(i)}$ by adding small perturbation to $\theta_{z_i}$:
\begin{align*}
\tilde{\mu}_{z_i}^{(i)} \sim \text{N}(\mu_{z_i},0.01), \quad 
\tilde{\delta}_{z_i}^{(i)} \sim \text{N}(\delta_{z_i},0.01), \quad
\tilde{\omega}_{z_i}^{(i)} \sim \text{N}(\omega_{z_i},0.05)
\end{align*}
The sequence is then sampled based on Hawkes process with parameter $\tilde{\theta}_{z_i}^{(i)}$ in time interval $[0,20]$.  To ease the tuning we normalize the sequences by dividing by the largest timestamp. 
We set $B_{k\ell} = \frac{0.5}{N}, \frac{1}{N}, \frac{2}{N}$, for any $k\neq \ell$, and $B_{kk}=\frac{5}{\# \{i \in [1,\cdots,N]: z_i=k\}}$. We sample the graph edges based on $\bm{B}$.  Denote $S = B_{k\ell} \times N$. The generated graphs  are visualized in the second column of Table \ref{tab:graphs}.

\textbf{Visualization of communities.}
We visualize the communities learned by HARMLESS (MAML) in Table \ref{tab:graphs}. Denote $K_0$ as the number of communities specified in HARMLESS.   We adopt $K_0$ colors corresponding to the $K_0$ communities in the graph. The color of each node shown in the Table \ref{tab:graphs} is the linear combinations of the RGB values of the $K_0$ colors weighted by identity proportions $\pi_i$.

HARMLESS produces reasonable identities even if $K_0$ is mis-specified.  If $K_0<K$, some of the communities would merge. If $K_0>K$, some of the communities would split.  


\begin{wrapfigure}{r}{0.45\textwidth}
\vspace{-25pt}
  \begin{center}
    \includegraphics[width=0.43\textwidth]{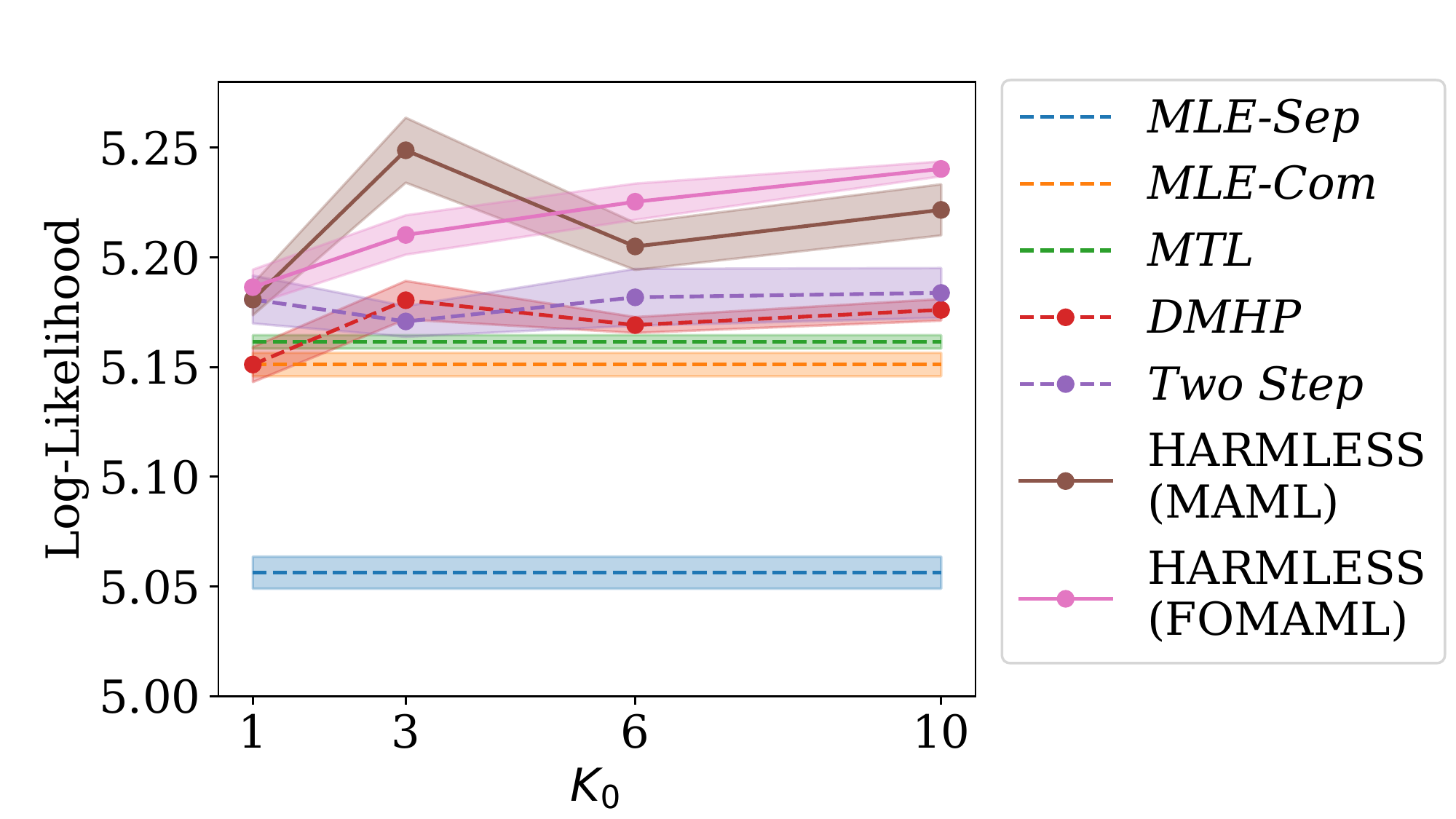}
  \end{center}
  \vspace{-10pt}
  \caption{\label{fig:synthetic} Plot of synthetic data. $S=1.$}
  \vspace{-15pt}
\end{wrapfigure}

\textbf{Benefit of joint training.}
To validate the benefit of joint training on graphs and sequences, we compare HARMLESS result with a two step procedure: We first train an MMB model and obtain the identities, and train HARMLESS (MAML) with fixed identities. In Figure \ref{fig:synthetic} we plot the obtained log-likelihood with respect to $K_0$. 

HARMLESS (MAML) consistently achieves larger log-likelihood than the two step procedure. This suggests joint training of graphs and the sequences indeed improve the prediction of future events. 

\textbf{Log-likelihood with respect to $K_0$.}
We also include the results of the baselines and HARMLESS (FOMAML) in Figure \ref{fig:synthetic}. The performance of HARMLESS is consistently better than the baselines. Besides, we find the performance HARMLESS (Reptile) is very dependent on the dataset. For this synthetic dataset, Reptile cannot perform well.






\begin{table}[]
\vspace{-10pt}
\caption{\label{tab:real} Log-likelihood of real datasets. }
\scalebox{0.88}{
\begin{tabular}{lcccc}
\hline
Dataset            & 911-Calls       &LinkedIn & MathOverflow & StackOverflow   \\ \hline
{\it MLE-Sep} &$4.0030\pm0.3763$ & $ 0.8419 \pm 0.0251$ & $0.5043\pm0.0657$&$0.2862\pm0.0177$\\ 
{\it MLE-Com} &$4.5111\pm0.3192$& $0.8768\pm 0.0028$ &$1.7805\pm0.0345$&$1.5594\pm0.0134$\\ 
{\it DMHP} & $4.4812\pm0.3434$& $0.8348 \pm 0.0030$&$1.5394\pm0.0347$& $N\backslash A$\\
{\it MTL} & $4.4621\pm0.3173$&$0.9270 \pm 0.0027$&$1.7225\pm0.0336$ &$1.4910\pm0.0089$\\
HARMLESS (MAML) &$4.5208\pm0.3256$& $\bm{1.4070}\pm 0.0105$ &$1.8563\pm0.0345$ &$1.3886\pm0.0082$\\
HARMLESS (FOMAML) &$\bm{4.6362}\pm 0.3241$ &$1.0129\pm 0.004$ &$1.8344\pm0.0348$&$1.5988\pm0.0083$\\ 
HARMLESS (Reptile) &$4.4929\pm0.3503$&  $0.9540\pm 0.0082$ &$\bm{1.8663}\pm0.0342$&$\bm{1.6017}\pm0.0097$\\ \hline
\end{tabular}}
\vspace{-15pt}
\end{table}

\vspace{-0.15in}
\subsection{Real Data}
\vspace{-0.1in}

We adopt four real datasets.

\textbf{911-Calls dataset}: The 911-Calls dataset\footnote{Data is provided by montcoalert.org.} contains emergency phone call records of fire, traffic and other emergencies for Montgomery County, PA. The county is divided into disjoint areas, each of which has a unique ZIP Code. For each area, the timestamps of emergency phone calls in this area are recorded as an event sequence. We consider each area as a subject, and two subjects are connected if they are adjoint. We finally obtain $57$ subjects and $81$ connections among them. The average length of the sequences is $219.1$. 


\textbf{LinkedIn dataset}: The LinkedIn dataset \citep{xu2017learning} contains job hopping records of the users. For each user, her/his check-in timestamps corresponding to different companies are recorded as an event sequence. We consider each user as a subject, and two subjects are connected if the difference in timestamps of two user joined the same company is less than 2 weeks. After removing the singleton subjects, we have $1,369$ subjects and $12,815$ connections among them. The average length of the sequences is $4.9$.


 \textbf{MathOverflow dataset}: The MathOverflow dataset \citep{paranjape2017motifs} contains  records of the users posting and answering math questions. We adopt the records from May 2, 2014 to March 6, 2016.
 For each user, her/his  timestamps of answering questions are recorded as an event sequence. We consider each user as a subject, and two subjects are connected if one user  answers another user's question. After removing the singleton subjects, we have $1,529$ subjects and $6,937$ connections among them. The average length of the sequences is $11.8$.

 
\textbf{StackOverflow dataset}: StackOverflow  is  a question and answer site similar to MathOverflow. We adopt the records from November 8, 2015 to December 1, 2015. We construct the sequences and graphs in the same way as MathOverflow. After removing the singleton subjects, we have $13,434$ users and $19,507$ connections among them.  The average length of the sequences is $7.7$.



\textbf{Result}: The log-likelihood is summarized in Table \ref{tab:real}. Note due to Markov chain Monte Carlo is needed for \textit{DMHP}, we cannot get reasonable result for large dataset, i.e., StackOverflow. HARMLESS performs consistently better than the baselines. Since the standard error of the results of 911-Calls dataset are large, we also performed a paired t test. The test shows the difference in log-likelihood between \textit{MLE-Com}, i.e., best of the baselines, and HARMLESS (FOMAML), i.e., best of HARMLESS series, is statistically significant (with $p$ value$=1.3\times 10^{-5}$).


\begin{wraptable}{R}{7.8cm}
\vspace{-30pt}
\caption{\label{tab:ablation} Results of ablation study.}
\vspace{-0.1in}
\scalebox{0.88}{
\begin{tabular}{ll}
\hline
Method                   &Log-Likelihood   \\ \hline
HARMLESS (MAML) & $ \bm{1.4070}\pm 0.0105$ \\ 
HARMLESS (FOMAML) & $ 1.0129\pm 0.0042$\\ 
HARMLESS (Reptile) & $0.9540\pm 0.0082$\\ \hline
Remove inner heterogeneity ($K=3$)& $0.9405 \pm 0.0032$\\ 
Remove inner heterogeneity ($K=5$)&  $0.9392 \pm 0.0032$ \\ 
Remove grouping (MAML) & $0.9432 \pm 0.0031$  \\
Remove grouping (FOMAML) & $0.9376 \pm 0.0031$ \\ 
Remove grouping (Reptile) &  $0.9455 \pm 0.0041$\\
Remove graph (MAML) & $0.9507 \pm 0.0032$  \\
Remove graph (FOMAML) &  $0.9446 \pm 0.0032$ \\ 
Remove graph (Reptile) & $0.9489 \pm 0.0072$ \\ \hline
\end{tabular}}
\vspace{-10pt}
\end{wraptable} 

\vspace{-0.17in}
\subsection{Ablation Study}
\vspace{-0.1in}

We then perform ablation study using LinkedIn dataset. Three sets of ablation study are considered here: \\
\textbf{Remove inner heterogeneity}: We model each community of sequences using the same parameters, i.e., we set $\tilde{\theta}_k^{(i)} = \theta_{k}$.  \\
\textbf{Remove grouping}: We set $K=1$, so that the whole graph is one community. This equivalent to apply the MAML-type algorithms on the sequences directly.\\
\textbf{Remove graph}: We do not consider the graph information, i.e., we remove $\bm{z}_{\to}$, $\bm{z}_{\leftarrow}$, $\bm{Y}$ and $\bm{B}$ from the panel in Figure \ref{fig:mmb+seq}. 

The results in Table \ref{tab:ablation} suggest that MAML-type adaptation, graph information, and using multiple identities all contribute to the good performance of HARMLESS.

\vspace{-0.2in}
\section{Discussions}
\vspace{-0.15in}

\textbf{The setting of meta learning.} The goal of conventional settings of meta learning is to train a model on a set of tasks, so that it can quickly adapt to a new task with only few training samples. Therefore, people divide the tasks into meta training set and meta test set, where each of the task contains a training set and a test set. The meta model is trained on the meta training set, aiming to minimize the test errors, and validated on the meta test set \citep{vinyals2016matching, santoro2016meta}. This setting is designed for supervised learning or reinforcement learning tasks that has accuracy or reward as a clear evaluation metric. Extracting information from the event sequences, however, is essentially an unsupervised learning task. Therefore, we do not separate meta training set and meta test set. Instead, we pull the collection of tasks together, and aim to extract shared information of the collection to help the training of models on individual tasks. Here, each short sequence is a task. We exploit the shared pattern of the collection of the sequences to obtain the models for individual sequences. 

\vspace{-0.02in}

\textbf{Community Pattern.} The target of Mixed Membership stochastic Blockmodels (MMB) is to identify the communities in a social graph, e.g., the classes in a school. However, real social graphs cannot always be viewed as Erd\H os-R\' enyi (ER) graphs assumed by MMB. As argued in  \citet{karrer2011stochastic}, for real-world networks, MMB tends to assign nodes with similar degrees to same communities, which is different from the popular interpretation of the community pattern. This property, however, is actually very helpful in our case. As an example, Twitter users that are more active tend to have similar behavior: They tend to make more connections and post tweets more frequently. In contrast, users with very different node degrees often have the tweets histories of different characteristics, and thus should be assigned to different identities. Such property of MMB allows the identities in HARMLESS to represent this non-traditional community patterns in non-ER graphs, i.e., it assigns subjects with various activeness to different communities.  

\vspace{-0.02in}

\textbf{Mixture of Hawkes processes.}  Many existing works adopt mixture of Hawkes process to model sequences that are generated from complicated mechanisms \citep{yang2013mixture, li2013dyadic, xu2017dirichlet}. Those works are different from HARMLESS since they do not consider the hierarchical heterogeneity of the sequences, and do not consider the relational information.

\vspace{-0.02in}

\textbf{Variants of Hawkes process.} Some attempts have been made to further enhance the flexibility of Hawkes processes. For example, the time-dependent Hawkes process (TiDeH) in \citet{kobayashi2016tideh} and the neural network-based Hawkes process (N-SM-MPP) in \citet{mei2017neural} learn very flexible Hawkes processes with complicated intensity functions. Those models usually have more parameters than vanilla Hawkes processes. 
For longer sequences, HARMLESS can also be naturally extended to TiDeHs or N-SM-MPP.
However, this work focuses on short sequences. These methods are not useful here, since they have too many degrees of freedom. 




\bibliographystyle{ims}
\bibliography{references}

\begin{thebibliography}{55}
\expandafter\ifx\csname natexlab\endcsname\relax\def\natexlab#1{#1}\fi
\expandafter\ifx\csname url\endcsname\relax
  \def\url#1{\texttt{#1}}\fi
\expandafter\ifx\csname urlprefix\endcsname\relax\def\urlprefix{}\fi

\bibitem[{Achab et~al.(2017)Achab, Bacry, Ga{\"\i}ffas, Mastromatteo and
  Muzy}]{achab2017uncovering}
\textsc{Achab, M.}, \textsc{Bacry, E.}, \textsc{Ga{\"\i}ffas, S.},
  \textsc{Mastromatteo, I.} and \textsc{Muzy, J.-F.} (2017).
\newblock Uncovering causality from multivariate hawkes integrated cumulants.
\newblock \textit{The Journal of Machine Learning Research}, \textbf{18}
  6998--7025.

\bibitem[{Airoldi et~al.(2008)Airoldi, Blei, Fienberg and
  Xing}]{airoldi2008mixed}
\textsc{Airoldi, E.~M.}, \textsc{Blei, D.~M.}, \textsc{Fienberg, S.~E.} and
  \textsc{Xing, E.~P.} (2008).
\newblock Mixed membership stochastic blockmodels.
\newblock \textit{Journal of machine learning research}, \textbf{9} 1981--2014.

\bibitem[{Bacry et~al.(2012)Bacry, Dayri and Muzy}]{bacry2012non}
\textsc{Bacry, E.}, \textsc{Dayri, K.} and \textsc{Muzy, J.-F.} (2012).
\newblock Non-parametric kernel estimation for symmetric hawkes processes.
  application to high frequency financial data.
\newblock \textit{The European Physical Journal B}, \textbf{85} 157.

\bibitem[{Bauwens and Hautsch(2009)}]{bauwens2009modelling}
\textsc{Bauwens, L.} and \textsc{Hautsch, N.} (2009).
\newblock Modelling financial high frequency data using point processes.
\newblock In \textit{Handbook of financial time series}. Springer, 953--979.

\bibitem[{Bengio et~al.(1990)Bengio, Bengio and Cloutier}]{bengio1990learning}
\textsc{Bengio, Y.}, \textsc{Bengio, S.} and \textsc{Cloutier, J.} (1990).
\newblock \textit{Learning a synaptic learning rule}.
\newblock Universit{\'e} de Montr{\'e}al, D{\'e}partement d'informatique et de
  recherche~?

\bibitem[{Blei et~al.(2017)Blei, Kucukelbir and
  McAuliffe}]{blei2017variational}
\textsc{Blei, D.~M.}, \textsc{Kucukelbir, A.} and \textsc{McAuliffe, J.~D.}
  (2017).
\newblock Variational inference: A review for statisticians.
\newblock \textit{Journal of the American Statistical Association},
  \textbf{112} 859--877.

\bibitem[{Blundell et~al.(2012)Blundell, Beck and
  Heller}]{blundell2012modelling}
\textsc{Blundell, C.}, \textsc{Beck, J.} and \textsc{Heller, K.~A.} (2012).
\newblock Modelling reciprocating relationships with hawkes processes.
\newblock In \textit{Advances in Neural Information Processing Systems}.

\bibitem[{Box and Tiao(2011)}]{box2011bayesian}
\textsc{Box, G.~E.} and \textsc{Tiao, G.~C.} (2011).
\newblock \textit{Bayesian inference in statistical analysis}, vol.~40.
\newblock John Wiley \& Sons.

\bibitem[{Chalmers(1991)}]{chalmers1991evolution}
\textsc{Chalmers, D.~J.} (1991).
\newblock The evolution of learning: An experiment in genetic connectionism.
\newblock In \textit{Connectionist Models}. Elsevier, 81--90.

\bibitem[{Cleeremans and McClelland(1991)}]{cleeremans1991learning}
\textsc{Cleeremans, A.} and \textsc{McClelland, J.~L.} (1991).
\newblock Learning the structure of event sequences.
\newblock \textit{Journal of Experimental Psychology: General}, \textbf{120}
  235.

\bibitem[{Eichler et~al.(2017)Eichler, Dahlhaus and
  Dueck}]{eichler2017graphical}
\textsc{Eichler, M.}, \textsc{Dahlhaus, R.} and \textsc{Dueck, J.} (2017).
\newblock Graphical modeling for multivariate hawkes processes with
  nonparametric link functions.
\newblock \textit{Journal of Time Series Analysis}, \textbf{38} 225--242.

\bibitem[{Farajtabar et~al.(2017)Farajtabar, Yang, Ye, Xu, Trivedi, Khalil, Li,
  Song and Zha}]{farajtabar2017fake}
\textsc{Farajtabar, M.}, \textsc{Yang, J.}, \textsc{Ye, X.}, \textsc{Xu, H.},
  \textsc{Trivedi, R.}, \textsc{Khalil, E.}, \textsc{Li, S.}, \textsc{Song, L.}
  and \textsc{Zha, H.} (2017).
\newblock Fake news mitigation via point process based intervention.
\newblock In \textit{Proceedings of the 34th International Conference on
  Machine Learning-Volume 70}. JMLR. org.

\bibitem[{Farajtabar et~al.(2016)Farajtabar, Ye, Harati, Song and
  Zha}]{farajtabar2016multistage}
\textsc{Farajtabar, M.}, \textsc{Ye, X.}, \textsc{Harati, S.}, \textsc{Song,
  L.} and \textsc{Zha, H.} (2016).
\newblock Multistage campaigning in social networks.
\newblock In \textit{Advances in Neural Information Processing Systems}.

\bibitem[{Finn et~al.(2017)Finn, Abbeel and Levine}]{finn2017model}
\textsc{Finn, C.}, \textsc{Abbeel, P.} and \textsc{Levine, S.} (2017).
\newblock Model-agnostic meta-learning for fast adaptation of deep networks.
\newblock In \textit{Proceedings of the 34th International Conference on
  Machine Learning-Volume 70}. JMLR. org.

\bibitem[{Finn et~al.(2018)Finn, Xu and Levine}]{finn2018probabilistic}
\textsc{Finn, C.}, \textsc{Xu, K.} and \textsc{Levine, S.} (2018).
\newblock Probabilistic model-agnostic meta-learning.
\newblock In \textit{Advances in Neural Information Processing Systems}.

\bibitem[{Fox et~al.(2016)Fox, Short, Schoenberg, Coronges and
  Bertozzi}]{fox2016modeling}
\textsc{Fox, E.~W.}, \textsc{Short, M.~B.}, \textsc{Schoenberg, F.~P.},
  \textsc{Coronges, K.~D.} and \textsc{Bertozzi, A.~L.} (2016).
\newblock Modeling e-mail networks and inferring leadership using self-exciting
  point processes.
\newblock \textit{Journal of the American Statistical Association},
  \textbf{111} 564--584.

\bibitem[{Girvan and Newman(2002)}]{girvan2002community}
\textsc{Girvan, M.} and \textsc{Newman, M.~E.} (2002).
\newblock Community structure in social and biological networks.
\newblock \textit{Proceedings of the national academy of sciences}, \textbf{99}
  7821--7826.

\bibitem[{Grant et~al.(2018)Grant, Finn, Levine, Darrell and
  Griffiths}]{grant2018recasting}
\textsc{Grant, E.}, \textsc{Finn, C.}, \textsc{Levine, S.}, \textsc{Darrell,
  T.} and \textsc{Griffiths, T.} (2018).
\newblock Recasting gradient-based meta-learning as hierarchical bayes.
\newblock \textit{arXiv preprint arXiv:1801.08930}.

\bibitem[{Hansen et~al.(2015)Hansen, Reynaud-Bouret, Rivoirard
  et~al.}]{hansen2015lasso}
\textsc{Hansen, N.~R.}, \textsc{Reynaud-Bouret, P.}, \textsc{Rivoirard, V.}
  \textsc{et~al.} (2015).
\newblock Lasso and probabilistic inequalities for multivariate point
  processes.
\newblock \textit{Bernoulli}, \textbf{21} 83--143.

\bibitem[{Hawkes(1971)}]{hawkes1971spectra}
\textsc{Hawkes, A.~G.} (1971).
\newblock Spectra of some self-exciting and mutually exciting point processes.
\newblock \textit{Biometrika}, \textbf{58} 83--90.

\bibitem[{Hoffman et~al.(2013)Hoffman, Blei, Wang and
  Paisley}]{hoffman2013stochastic}
\textsc{Hoffman, M.~D.}, \textsc{Blei, D.~M.}, \textsc{Wang, C.} and
  \textsc{Paisley, J.} (2013).
\newblock Stochastic variational inference.
\newblock \textit{The Journal of Machine Learning Research}, \textbf{14}
  1303--1347.

\bibitem[{Karrer and Newman(2011)}]{karrer2011stochastic}
\textsc{Karrer, B.} and \textsc{Newman, M.~E.} (2011).
\newblock Stochastic blockmodels and community structure in networks.
\newblock \textit{Physical review E}, \textbf{83} 016107.

\bibitem[{Kobayashi and Lambiotte(2016)}]{kobayashi2016tideh}
\textsc{Kobayashi, R.} and \textsc{Lambiotte, R.} (2016).
\newblock Tideh: Time-dependent hawkes process for predicting retweet dynamics.
\newblock In \textit{Tenth International AAAI Conference on Web and Social
  Media}.

\bibitem[{Koch et~al.(2015)Koch, Zemel and Salakhutdinov}]{koch2015siamese}
\textsc{Koch, G.}, \textsc{Zemel, R.} and \textsc{Salakhutdinov, R.} (2015).
\newblock Siamese neural networks for one-shot image recognition.
\newblock In \textit{ICML deep learning workshop}, vol.~2.

\bibitem[{Laub et~al.(2015)Laub, Taimre and Pollett}]{laub2015hawkes}
\textsc{Laub, P.~J.}, \textsc{Taimre, T.} and \textsc{Pollett, P.~K.} (2015).
\newblock Hawkes processes.
\newblock \textit{arXiv preprint arXiv:1507.02822}.

\bibitem[{Li and Zha(2013)}]{li2013dyadic}
\textsc{Li, L.} and \textsc{Zha, H.} (2013).
\newblock Dyadic event attribution in social networks with mixtures of hawkes
  processes.
\newblock In \textit{Proceedings of the 22nd ACM international conference on
  Information \& Knowledge Management}. ACM.

\bibitem[{Linderman and Adams(2014)}]{linderman2014discovering}
\textsc{Linderman, S.} and \textsc{Adams, R.} (2014).
\newblock Discovering latent network structure in point process data.
\newblock In \textit{International Conference on Machine Learning}.

\bibitem[{Luo et~al.(2015)Luo, Xu, Zhen, Ning, Zha, Yang and
  Zhang}]{luo2015multi}
\textsc{Luo, D.}, \textsc{Xu, H.}, \textsc{Zhen, Y.}, \textsc{Ning, X.},
  \textsc{Zha, H.}, \textsc{Yang, X.} and \textsc{Zhang, W.} (2015).
\newblock Multi-task multi-dimensional hawkes processes for modeling event
  sequences.
\newblock In \textit{Twenty-Fourth International Joint Conference on Artificial
  Intelligence}.

\bibitem[{Maclaurin et~al.(2015)Maclaurin, Duvenaud and
  Adams}]{maclaurin2015gradient}
\textsc{Maclaurin, D.}, \textsc{Duvenaud, D.} and \textsc{Adams, R.} (2015).
\newblock Gradient-based hyperparameter optimization through reversible
  learning.
\newblock In \textit{International Conference on Machine Learning}.

\bibitem[{Mei and Eisner(2017)}]{mei2017neural}
\textsc{Mei, H.} and \textsc{Eisner, J.~M.} (2017).
\newblock The neural hawkes process: A neurally self-modulating multivariate
  point process.
\newblock In \textit{Advances in Neural Information Processing Systems}.

\bibitem[{Munkhdalai and Yu(2017)}]{munkhdalai2017meta}
\textsc{Munkhdalai, T.} and \textsc{Yu, H.} (2017).
\newblock Meta networks.
\newblock In \textit{Proceedings of the 34th International Conference on
  Machine Learning-Volume 70}. JMLR. org.

\bibitem[{Nichol et~al.(2018)Nichol, Achiam and Schulman}]{nichol2018first}
\textsc{Nichol, A.}, \textsc{Achiam, J.} and \textsc{Schulman, J.} (2018).
\newblock On first-order meta-learning algorithms.
\newblock \textit{arXiv preprint arXiv:1803.02999}.

\bibitem[{Nichol and Schulman(2018)}]{nichol2018reptile}
\textsc{Nichol, A.} and \textsc{Schulman, J.} (2018).
\newblock Reptile: a scalable metalearning algorithm.
\newblock \textit{arXiv preprint arXiv:1803.02999}.

\bibitem[{Ogata(1999)}]{ogata1999seismicity}
\textsc{Ogata, Y.} (1999).
\newblock Seismicity analysis through point-process modeling: A review.
\newblock In \textit{Seismicity patterns, their statistical significance and
  physical meaning}. Springer, 471--507.

\bibitem[{Paranjape et~al.(2017)Paranjape, Benson and
  Leskovec}]{paranjape2017motifs}
\textsc{Paranjape, A.}, \textsc{Benson, A.~R.} and \textsc{Leskovec, J.}
  (2017).
\newblock Motifs in temporal networks.
\newblock In \textit{Proceedings of the Tenth ACM International Conference on
  Web Search and Data Mining}. ACM.

\bibitem[{Rasmussen(2013)}]{rasmussen2013bayesian}
\textsc{Rasmussen, J.~G.} (2013).
\newblock Bayesian inference for hawkes processes.
\newblock \textit{Methodology and Computing in Applied Probability},
  \textbf{15} 623--642.

\bibitem[{Ravi and Beatson(2018)}]{ravi2018amortized}
\textsc{Ravi, S.} and \textsc{Beatson, A.} (2018).
\newblock Amortized bayesian meta-learning.

\bibitem[{Ravi and Larochelle(2016)}]{ravi2016optimization}
\textsc{Ravi, S.} and \textsc{Larochelle, H.} (2016).
\newblock Optimization as a model for few-shot learning.

\bibitem[{Reynaud-Bouret et~al.(2010)Reynaud-Bouret, Schbath
  et~al.}]{reynaud2010adaptive}
\textsc{Reynaud-Bouret, P.}, \textsc{Schbath, S.} \textsc{et~al.} (2010).
\newblock Adaptive estimation for hawkes processes; application to genome
  analysis.
\newblock \textit{The Annals of Statistics}, \textbf{38} 2781--2822.

\bibitem[{Ross et~al.(1996)Ross, Kelly, Sullivan, Perry, Mercer, Davis,
  Washburn, Sager, Boyce and Bristow}]{ross1996stochastic}
\textsc{Ross, S.~M.}, \textsc{Kelly, J.~J.}, \textsc{Sullivan, R.~J.},
  \textsc{Perry, W.~J.}, \textsc{Mercer, D.}, \textsc{Davis, R.~M.},
  \textsc{Washburn, T.~D.}, \textsc{Sager, E.~V.}, \textsc{Boyce, J.~B.} and
  \textsc{Bristow, V.~L.} (1996).
\newblock \textit{Stochastic processes}, vol.~2.
\newblock Wiley New York.

\bibitem[{Santoro et~al.(2016)Santoro, Bartunov, Botvinick, Wierstra and
  Lillicrap}]{santoro2016meta}
\textsc{Santoro, A.}, \textsc{Bartunov, S.}, \textsc{Botvinick, M.},
  \textsc{Wierstra, D.} and \textsc{Lillicrap, T.} (2016).
\newblock Meta-learning with memory-augmented neural networks.
\newblock In \textit{International conference on machine learning}.

\bibitem[{Snell et~al.(2017)Snell, Swersky and Zemel}]{snell2017prototypical}
\textsc{Snell, J.}, \textsc{Swersky, K.} and \textsc{Zemel, R.} (2017).
\newblock Prototypical networks for few-shot learning.
\newblock In \textit{Advances in Neural Information Processing Systems}.

\bibitem[{Sung et~al.(2018)Sung, Yang, Zhang, Xiang, Torr and
  Hospedales}]{sung2018learning}
\textsc{Sung, F.}, \textsc{Yang, Y.}, \textsc{Zhang, L.}, \textsc{Xiang, T.},
  \textsc{Torr, P.~H.} and \textsc{Hospedales, T.~M.} (2018).
\newblock Learning to compare: Relation network for few-shot learning.
\newblock In \textit{Proceedings of the IEEE Conference on Computer Vision and
  Pattern Recognition}.

\bibitem[{Tran et~al.(2015)Tran, Farajtabar, Song and Zha}]{tran2015netcodec}
\textsc{Tran, L.}, \textsc{Farajtabar, M.}, \textsc{Song, L.} and \textsc{Zha,
  H.} (2015).
\newblock Netcodec: Community detection from individual activities.
\newblock In \textit{Proceedings of the 2015 SIAM International Conference on
  Data Mining}. SIAM.

\bibitem[{Trivedi et~al.(2018)Trivedi, Farajtabar, Biswal and
  Zha}]{trivedi2018dyrep}
\textsc{Trivedi, R.}, \textsc{Farajtabar, M.}, \textsc{Biswal, P.} and
  \textsc{Zha, H.} (2018).
\newblock Dyrep: Learning representations over dynamic graphs.

\bibitem[{Vinyals et~al.(2016)Vinyals, Blundell, Lillicrap, Wierstra
  et~al.}]{vinyals2016matching}
\textsc{Vinyals, O.}, \textsc{Blundell, C.}, \textsc{Lillicrap, T.},
  \textsc{Wierstra, D.} \textsc{et~al.} (2016).
\newblock Matching networks for one shot learning.
\newblock In \textit{Advances in neural information processing systems}.

\bibitem[{Xie et~al.(2013)Xie, Kelley and Szymanski}]{xie2013overlapping}
\textsc{Xie, J.}, \textsc{Kelley, S.} and \textsc{Szymanski, B.~K.} (2013).
\newblock Overlapping community detection in networks: The state-of-the-art and
  comparative study.
\newblock \textit{Acm computing surveys (csur)}, \textbf{45} 43.

\bibitem[{Xu et~al.(2017{\natexlab{a}})Xu, Luo, Chen and
  Carin}]{xu2017benefits}
\textsc{Xu, H.}, \textsc{Luo, D.}, \textsc{Chen, X.} and \textsc{Carin, L.}
  (2017{\natexlab{a}}).
\newblock Benefits from superposed hawkes processes.
\newblock \textit{arXiv preprint arXiv:1710.05115}.

\bibitem[{Xu et~al.(2017{\natexlab{b}})Xu, Luo and Zha}]{xu2017learning}
\textsc{Xu, H.}, \textsc{Luo, D.} and \textsc{Zha, H.} (2017{\natexlab{b}}).
\newblock Learning hawkes processes from short doubly-censored event sequences.
\newblock In \textit{Proceedings of the 34th International Conference on
  Machine Learning-Volume 70}. JMLR. org.

\bibitem[{Xu and Zha(2017)}]{xu2017dirichlet}
\textsc{Xu, H.} and \textsc{Zha, H.} (2017).
\newblock A dirichlet mixture model of hawkes processes for event sequence
  clustering.
\newblock In \textit{Advances in Neural Information Processing Systems}.

\bibitem[{Yang and Zha(2013)}]{yang2013mixture}
\textsc{Yang, S.-H.} and \textsc{Zha, H.} (2013).
\newblock Mixture of mutually exciting processes for viral diffusion.
\newblock In \textit{International Conference on Machine Learning}.

\bibitem[{Zarezade et~al.(2017)Zarezade, Khodadadi, Farajtabar, Rabiee and
  Zha}]{zarezade2017correlated}
\textsc{Zarezade, A.}, \textsc{Khodadadi, A.}, \textsc{Farajtabar, M.},
  \textsc{Rabiee, H.~R.} and \textsc{Zha, H.} (2017).
\newblock Correlated cascades: Compete or cooperate.
\newblock In \textit{Thirty-First AAAI Conference on Artificial Intelligence}.

\bibitem[{Zhang and Yang(2017)}]{zhang2017survey}
\textsc{Zhang, Y.} and \textsc{Yang, Q.} (2017).
\newblock A survey on multi-task learning.
\newblock \textit{arXiv preprint arXiv:1707.08114}.

\bibitem[{Zhao et~al.(2015)Zhao, Erdogdu, He, Rajaraman and
  Leskovec}]{zhao2015seismic}
\textsc{Zhao, Q.}, \textsc{Erdogdu, M.~A.}, \textsc{He, H.~Y.},
  \textsc{Rajaraman, A.} and \textsc{Leskovec, J.} (2015).
\newblock Seismic: A self-exciting point process model for predicting tweet
  popularity.
\newblock In \textit{Proceedings of the 21th ACM SIGKDD International
  Conference on Knowledge Discovery and Data Mining}. ACM.

\bibitem[{Zhou et~al.(2013)Zhou, Zha and Song}]{zhou2013learning}
\textsc{Zhou, K.}, \textsc{Zha, H.} and \textsc{Song, L.} (2013).
\newblock Learning social infectivity in sparse low-rank networks using
  multi-dimensional hawkes processes.
\newblock In \textit{Artificial Intelligence and Statistics}.

\end{thebibliography}
\appendix
\newpage
\section{Related Works}

\textbf{Hawkes Process}  Hawkes process has long been used to model event sequences \citep{hawkes1971spectra}, such as earthquake aftershock sequences \citep{ogata1999seismicity}, financial transactions \citep{bauwens2009modelling}, and events on social networks \citep{fox2016modeling, farajtabar2017fake}. Its variant, mixture of Hawkes processes model, has also been proved effective in many area \citep{yang2013mixture, li2013dyadic, xu2017dirichlet}. 
In most cases, the learning methodology is variational inference or maximum likelihood estimation \citep{rasmussen2013bayesian,zhou2013learning,zhao2015seismic}. Other possible methods includes least-squares-based method \citep{eichler2017graphical}, Wiener-Hopf-based methods \citep{bacry2012non}, and cumulants-based methods \citep{achab2017uncovering}. 

Instead of predefine an impact function here, some non-parametric methods use discretization or kernel-estimation when learning models \citep{reynaud2010adaptive, zhou2013learning, hansen2015lasso}. Those methods usually target small datasets, and do not need a good scalability. Recently, some attempts have been made to further enhance the flexibility of Hawkes processes. The time-dependent Hawkes process (TiDeH) in \citet{kobayashi2016tideh} and the neural network-based Hawkes process in \citet{mei2017neural} learn very flexible Hawkes processes with complicated intensity functions. Those methods usually target very long and multi-dimensional sequences, instead of short sequences.

Existing works targeting short sequences is usually in specific cases \citep{xu2017benefits,xu2017learning}, such as the data is censored. However, there is no work targeting general short sequences as we do here.

There are lines of research that involves both point processes and graphs. One is using point process to find the latent graph \citep{blundell2012modelling, linderman2014discovering, tran2015netcodec}. Another one is considering the interaction of the nodes as point process and use it to construct a dynamic graph, instead of the event happens on nodes as we consider here \citep{farajtabar2016multistage, zarezade2017correlated,trivedi2018dyrep}. These works have vary different aims from our work.

\textbf{Meta Learning} Meta learning has been studied since last century \citep{bengio1990learning, chalmers1991evolution}. Some works focus on learning the hyperparameters, such as learning rates or initial conditions \citep{maclaurin2015gradient}. Some works aim to learn a metric so that a simple K nearest neighbors can perform well under such a metric \citep{koch2015siamese, vinyals2016matching, sung2018learning, snell2017prototypical}. Some works design specific deep neural networks so that the information of different tasks are memorized and thus the model can easily generalize to new tasks \citep{santoro2016meta, munkhdalai2017meta,ravi2016optimization}. 

Model-Agnostic Meta Learning (MAML) method \citep{finn2017model} opens another line of research, i.e., it designs an optimization scheme so that the model can fast adapt to new tasks. Reptile \citep{nichol2018reptile}, a variant of MAML, is proposed to simplify the computation of MAML. None of those works, however, considers the relational information between tasks like our method, which is critical in modeling short sequences.

One interesting line of follow-up works of MAML is connecting MAML with Bayesian inference \citep{finn2018probabilistic, ravi2018amortized, grant2018recasting}. Since HARMLESS combines a Bayesian model with MAML, it has the potential to be rewritten into a pure Bayesian model that has better quantification of uncertainty. We left this for future work.  



\section{Definition of Operator $\mathcal{D}$}
\label{sec:defD}

As we mentioned earlier, 
\begin{align*}
\min_{\theta} \sum_{\mathcal{T}_i\in \Gamma} \mathcal{F}_{\mathcal{T}_i} (\tilde{\theta}_i) = \sum_{\mathcal{T}_i\in \Gamma} \mathcal{F}_{\mathcal{T}_i} (\theta-\eta \mathcal{D}( \mathcal{F}_{\mathcal{T}_i}, \theta))
\end{align*}
is the loss function for MAML, FOMAML, and Reptile algorithm with different definition of the operator $\mathcal{D}$.

For simplicity, here we define the operator of one gradient step. The cases of few gradient steps can be defined analogously. 

For MAML, $\mathcal{D}(\mathcal{F}_{\mathcal{T}_i}, \theta)$ is defined as $\nabla_{\theta}(\mathcal{F}_{\mathcal{T}_i}(\theta))$.

For First Order MAML (FOMAML), $\mathcal{D}(\mathcal{F}_{\mathcal{T}_i}, \theta)$ is also defined as $\nabla_{\theta}(\mathcal{F}_{\mathcal{T}_i}(\theta))$. The difference is that the output of the operator just a value, not a function of $\theta$, i.e., when we solve the gradient of $\mathcal{F}_{\mathcal{T}_i} (\theta-\eta \mathcal{D}( \mathcal{F}_{\mathcal{T}_i}, \theta))$, the gradient does not back-propagate into $\mathcal{D}(\mathcal{F}_{\mathcal{T}_i}, \theta)$.

For Reptile, the algorithm of reptile is as follows \cite{nichol2018reptile}.

\begin{algorithm}[H]
 \caption{Reptile}
 \begin{algorithmic}
 \WHILE{not converged}
 \STATE Sample task $\mathcal{T}$ with loss $\mathcal{F}_{\mathcal{T}}$;
 \STATE $W \leftarrow \text{SGD}(\mathcal{F}_{\mathcal{T}} , \theta, k)$, where $k$ is the number of SGD steps; 
 \STATE Do the update $\theta \leftarrow \theta - \eta(\theta-W)$;
 \ENDWHILE
\end{algorithmic}
\end{algorithm}
From the algorithm we can see, operator $\mathcal{D}$ is defined as $\mathcal{D}(\mathcal{F}_{\mathcal{T}}, \theta) = \text{SGD}(\mathcal{F}_{\mathcal{T}} , \theta, 1)$. Similar as FOMAML, computing the gradient also does not back-propagate into $\mathcal{D}(\mathcal{F}_{\mathcal{T}_i}, \theta)$.

\section{Derivation of Variational EM}
\label{sec:derEM}

\textbf{Preparation}
After adding latent variable $\bm{z}$, the joint distribution is
\begin{align*}
    p(\bm{T},\bm{Y}, \bm{z} ,\bm{z}_{\rightarrow},\bm{z}_{\leftarrow}, \bm{\pi}) &= p(\bm{T}| \bm{z} )p(\bm{Y}|\bm{z}_{\rightarrow},\bm{z}_{\leftarrow})p(\bm{z}|\bm{\pi})  p(\bm{z}_{\leftarrow}|\bm{\pi})p(\bm{z}_{\rightarrow}|\bm{\pi})p(\bm{\pi}).
\end{align*}
where
\begin{align*}
    & p(\bm{T}|  \bm{z} ) = \prod_{i=1}^N \prod_{k=1}^K \left(\mathcal{L}_i(\theta_k -\eta \mathcal{D}( \mathcal{L}_i,\theta_k))\right)^{z_{i,k}}, \\
    & p(\bm{Y}|\bm{z}_{\rightarrow},\bm{z}_{\leftarrow}) = \prod_{i=1}^N \prod_{j=1}^N (z_{i\to j}^T \bm{B} z_{i \leftarrow j})^{Y_{ij}} (1-z_{i\to j}^T \bm{B} z_{i \leftarrow j})^{1-Y_{ij}}\\
    & p(\bm{z}|\bm{\pi}) = \prod_{i=1}^N \prod_{k=1}^K \pi_{i,k}^{z_{i,k}},\\
    & p(\bm{z}_{\to}|\bm{\pi}) = \prod_{i=1}^N \prod_{j=1}^N \prod_{k=1}^K \pi_{i,k}^{z_{i\to j,k}},\\
    & p(\bm{z}_{\leftarrow}|\bm{\pi}) = \prod_{i=1}^N \prod_{j=1}^N \prod_{k=1}^K \pi_{j,k}^{z_{i\leftarrow j,k}},\\
    & p(\bm{\pi}) = \prod_{i=1}^N \text{Dirichlet}(\pi_i|\alpha) = \prod_{i=1}^N C(\alpha) \prod_{k=1}^K \pi_{i,k}^{\alpha-1}.
\end{align*}
Note that in this section we represent $z_i, z_{i\to j}, z_{i\leftarrow j}$ as one-hot vector, while in the main paper we use scalar $z_i=k$ representing the identities.

The posterior distribution is defined as
\begin{align*}
p( \bm{z} ,\bm{z}_{\rightarrow},\bm{z}_{\leftarrow}, \bm{\pi} | \bm{T},\bm{Y},\alpha, \bm{\theta}, B).
\end{align*}

We aim to find a distribution $q(\bm{z} , \bm{z}_{\rightarrow},\bm{z}_{\leftarrow}, \bm{\pi})\in Q$, such that the Kullback-Leibler (KL) divergence between the above posterior distribution and $q(\bm{z} , \bm{z}_{\rightarrow},\bm{z}_{\leftarrow}, \bm{\pi})$ is minimized. This can be achieved by maximize the Evidence Lower BOund (ELBO),
\begin{align*}
    \mathcal{B}(q) = \mathbb{E}_q [\log p( \bm{z} ,\bm{z}_{\rightarrow},\bm{z}_{\leftarrow}, \bm{\pi}, \bm{T},\bm{Y})] - \mathbb{E}_q[\log q(\bm{z} , \bm{z}_{\rightarrow},\bm{z}_{\leftarrow}, \bm{\pi})].
\end{align*}

\textbf{Variational family} We adopt the mean-field variational family, i.e.,
\begin{align*}
    q(\bm{z} , \bm{z}_{\rightarrow},\bm{z}_{\leftarrow}, \bm{\pi}) =  q_1(\bm{\pi})\prod_i q_2(z_i)\prod_j q_3(z_{i\to j}) q_4(z_{i\leftarrow j}).
\end{align*}
We pick $q_1(\pi_i)$ as PDF of $\text{Dirichlet}(\beta)$, $q_2(z_i)$ as PDF of $\text{Categorical}(\gamma_i)$, $q_3(z_{i\to j})$ as PDF of $\text{Categorical}(\phi_{ij})$, $q_4(z_{i\leftarrow j})$ as PDF of $\text{Categorical}(\psi_{ij})$.

\textbf{Update for $q_1$} Again, our goal is to maximize  
\begin{align*}
    \mathcal{B}(q) = \mathbb{E}_q [\log p( \bm{z} ,\bm{z}_{\rightarrow},\bm{z}_{\leftarrow}, \bm{\pi}, \bm{T},\bm{Y})] - \mathbb{E}_q[\log q(\bm{z} , \bm{z}_{\rightarrow},\bm{z}_{\leftarrow}, \bm{\pi})].
\end{align*}
Now we focus on $q_1$, and treat $q_2$, $q_3$ and $q_4$ as given. We want to maximize
\begin{align*}
    \mathcal{F}_{\bm{\pi}}(q_1) &= \mathbb{E}_q [\log p( \bm{z} ,\bm{z}_{\rightarrow},\bm{z}_{\leftarrow}, \bm{\pi}, \bm{T},\bm{Y})] - \mathbb{E}_q[\log q(\bm{z} , \bm{z}_{\rightarrow},\bm{z}_{\leftarrow}, \bm{\pi})]\\
    & = \mathbb{E}_q [\log p(\bm{T}|  \bm{z} ) + \log p(\bm{Y}|\bm{z}_{\leftarrow}, \bm{z}_{\to})+ \log p(\bm{z}|\bm{\pi}) + \log p(\bm{z}_{\leftarrow}| \bm{\pi})+\log p(\bm{z}_{\to}|\bm{\pi})+\log p(\bm{\pi})] \\
    & \qquad - \mathbb{E}_{q_1}[\log q_1(\bm{\pi})] + \text{const}\\
    & = \mathbb{E}_q [ \log p(\bm{z}|\bm{\pi})+ \log p(\bm{z}_{\leftarrow}| \bm{\pi})+\log p(\bm{z}_{\to}|\bm{\pi})+\log p(\bm{\pi})] - \mathbb{E}_{q_1}[\log q_1(\bm{\pi})] + \text{const}\\
    & = \int q_1(\bm{\pi}) \left(\mathbb{E}_{q_2} [ \log p(\bm{z}|\bm{\pi})+ \log p(\bm{z}_{\leftarrow}| \bm{\pi})+\log p(\bm{z}_{\to}|\bm{\pi})+\log p(\bm{\pi})] -\log q_1(\bm{\pi}) \right) d\bm{\pi}+ \text{const}.
\end{align*}
Take the derivative, 
\begin{align*}
    \frac{\delta \mathcal{F}_{\bm{\pi}}(q_1)}{\delta q_1} = \mathbb{E}_{q_2} [ \log p(\bm{z}|\bm{\pi})+ \log p(\bm{z}_{\leftarrow}| \bm{\pi})+\log p(\bm{z}_{\to}|\bm{\pi})+\log p(\bm{\pi})] -\log q_1(\bm{\pi}) -1 =0.
\end{align*}
Substitute the expressions of the distributions, after some derivation we get the update for $\bm{\beta}$ as
\begin{align}
   \beta_{i,k} \leftarrow \alpha_k + \gamma_{i,k} + \sum_{j=1}^N \phi_{ij,k}+\sum_{j=1}^N \psi_{ij,k}. \label{eq:update_alpha1m}
\end{align}

\textbf{Update for $q_2$} Similarly, we have
\begin{align*}
    \mathcal{F}_{\bm{z}}(q_2) & = \mathbb{E}_q [ \log p(\bm{T}| \bm{z} ) +\log p(\bm{z}|\bm{\pi})] - \mathbb{E}_{q_2}[\log q_2(\bm{z})] + \text{const}\\
    & = \int q_2(\bm{z}) \left(\mathbb{E}_{q_1} [ \log p(\bm{T}| \bm{\theta}, \bm{z} ) +\log p(\bm{z}|\bm{\pi})] -\log q_2(\bm{z}) \right) d\bm{z}+ \text{const}.
\end{align*}
Take the derivative, 
\begin{align*}
    \frac{\delta \mathcal{F}_{\bm{z}}(q_2)}{\delta q_2} = \log p(\bm{T}| \bm{\theta}, \bm{z} ) + \mathbb{E}_{q_1} [ \log p(\bm{z}|\bm{\pi})] -\log q_2(\bm{z}) -1 = 0.
\end{align*}
After some derivation, we have
\begin{align}
    &\gamma_{i,k} \leftarrow \mathcal{L}_i(\theta_k -\eta \mathcal{D}( \mathcal{L}_i,\theta_k)) \exp \left( f_{\rm dg}(\beta_{i,k})-f_{\rm dg}(\sum_\ell \beta_{i,\ell}) \right), \label{eq:update_gamma1m}\\
    &\gamma_{i,k} \leftarrow \frac{\gamma_{i,k}}{\sum_{\ell} \gamma_{i,\ell}}, \label{eq:update_gamma2m}
\end{align}
where $f_{\rm dg}$ is the digamma function.

\textbf{Update for $q_3$ and $q_4$} The derivation of update for $q_3$ and $q_4$ is very similar to the update for $q_2$, so we will not elaborate on that. Readers who are interested might also refer to \cite{airoldi2008mixed}. The updates are
 \begin{align}
    & \phi_{ij,k} \leftarrow e^{\mathbb{E}_q [\log \pi_{i,k}]} \prod_{\ell=1}^K \left(B_{k\ell}^{Y_{ij}} (1-B_{k\ell})^{1-Y_{ij}}  \right)^{\psi_{ij,\ell}}, \quad \phi_{ij,k} \leftarrow \frac{\phi_{ij,k}}{\sum_{\ell} \phi_{ij,\ell}}, \label{eq:update_phi2_compm} \\
    & \psi_{ij,\ell} \leftarrow e^{\mathbb{E}_q [\log \pi_{j,\ell}]} \prod_{k=1}^K \left((B_{k\ell})^{Y_{ij}} (1-B_{k\ell})^{1-Y_{ij}}  \right)^{\phi_{ij,k}}, \quad \psi_{ij,k} \leftarrow \frac{\psi_{ij,k}}{\sum_{\ell} \psi_{ij,\ell}}, \label{eq:update_psi2_compm} 
    \end{align}

\textbf{Update for $\bm{\theta}$} We update $\bm{\theta}$ using gradient ascent. We first pick the terms that is relevant to $\bm{\theta}$, 
\begin{align*}
    \mathcal{F}_{\bm{\theta}}(\bm{\theta}) & = \mathbb{E}_q [ \log p(\bm{T}| \bm{\theta}, \bm{z} ) ]  + \text{const}\\
    & = \int q_2(\bm{z}) [ \log p(\bm{T}| \bm{\theta}, \bm{z} )]d \bm{z} + \text{const}\\
    & =\sum_{i=1}^N \sum_{k=1}^K \gamma_{i,k} \log \mathcal{L}_i(\theta_k -\eta \mathcal{D}( \mathcal{L}_i,\theta_k) )+ \text{const}.
\end{align*}
So the gradient ascent update is,
\begin{align}
    \bm{\theta} \leftarrow \bm{\theta} + \eta_1 \nabla_{\bm{\theta}} \left(\sum_{i=1}^N \sum_{k=1}^K \gamma_{i,k} \log \mathcal{L}_i(\theta_k -\eta \mathcal{D}( \mathcal{L}_i,\theta_k)) \right). \label{eq:update_kappam}
\end{align}

\textbf{Update for $\alpha$ and $B$} From \cite{airoldi2008mixed}, we have the update for $\alpha$ and $\bm{B}$ as follows

    \begin{align}
    & \alpha_{k} \leftarrow \alpha_{k} + 
    \eta_{\alpha} \left( N \big( f_{\rm dg}(\sum_\ell \alpha_{\ell}) - f_{\rm dg}(\alpha_{k})\big)+ \sum_{i=1}^N\big(  f_{\rm dg}(\beta_{i,k})-f_{\rm dg}(\sum_\ell \beta_{i,\ell})\big)\right), \label{eq:update_alpha_compm} \\
    & B_{k\ell} \leftarrow \frac{\sum_{ij} Y_{ij} \phi_{ij,k} \psi_{ij,\ell}}{\sum_{ij} \phi_{ij,k} \psi_{ij,\ell}}, \label{eq:update_B_compm}
    \end{align}

\section{Derivation of Evaluation Metric}
\label{sec:derEv}

In this section, we give more details on the evaluate metrics. 
Specifically, we show how to compute the NLL of the test set. 
Given a sequence $\bm{\tau}_i = \{\tau_i^{(1)}, \tau_i^{(2)}, \cdots, \tau_i^{(M_i)}\}$, we would like to predict the timestamp of $\tau_i^{(M_i+1)}$. Here, we use the probability of the arrival at time $\tau_i^{(M_i+1)}$ and no arrival in $[\tau_i^{(M_i)}, \tau_i^{(M_i+1)}]$ given history before $\tau_i^{(M_i)}$ as evaluation metric. 

Consider a Hawkes process with parameter $\theta$, the probability density is
\begin{align*}
    \mathcal{P} (\theta) & =  \lambda \big(\tau_i^{(M_i+1)}; \theta , \bm{\tau}_i)\big) \exp \Big( -\int_{\tau_i^{(M_i)}}^{\tau_i^{(M_i+1)}} \lambda(t ; \theta , \bm{\tau}_i) ~ dt\Big) \\ 
    & =\big(\mu + \sum_{m=1}^{M_i} \delta \omega e^{-\omega(\tau_i^{(M_i+1)}-\tau_i^{(m)})}\big) \exp \left(  -\mu (\tau_i^{(M_i+1)}-\tau_i^{(M_i)})  -\delta (1-e^{-\omega(\tau_i^{(M_i+1)}-\tau_i^{(M_i)})})\right).
\end{align*}

In the generative process, for subject $i$, we first sample $z_i$, then use parameter $\tilde{\theta}_{z_i}^{(i)} = \theta_{z_i} - \eta \mathcal{D}( \mathcal{L}_i,\theta_{z_i})$. The posterior distribution of $z_i$ is $q_2 (z_i), i.e., \text{Categorical}(\gamma_i)$. Therefore we have
\begin{align*}
    \mathbb{P}(z_i=k) = \gamma_{i,k}.
\end{align*}
So the likelihood of next arrival $\tau_i^{(M_i+1)}$ is
\begin{align*}
    \tilde{ \mathcal{L}}_{i}  & = \sum_{k=1}^K \mathbb{P}(z_i=k) \mathbb{P}(\text{next arrival is } \tau_i^{(M_i+1)}|\text{ Hawkes model with } \theta_k )\\
    & = \sum_{k=1}^K \gamma_{i,k} \mathcal{P}(\tilde{\theta}_k^{(i)} ).
\end{align*}
And then we sum $\tilde{ \mathcal{L}}_{i}$ over every subject.

%

\section{Detailed Settings of the Experiments}
\label{sec:set_exp}

Note that we can also adopt a non-informative $\alpha$ instead of updating it in every iteration. After some trial experiments, we find setting $\alpha = \bm{1}_K$ is numerically more stable than updating it in every iteration. Therefore we adopt $\alpha = \bm{1}_K$ in the following experiments. 

Besides, we find that $\nu$ causes nearly no effect to the result when varying from $10^{-10}$ to $10^{-1}$. We fix it as $10^{-2}$.

\subsection{Synthetic Dataset}
Both the baselines and our proposed methods are fine tuned. We first perform a coarse grid search to find hyper-parameters for all methods. The grid search finds learning rate from $1 \times 10^{-7}$ to $1$ for both inner and outer updates.  To perform the multi-split procedure, all hyper-parameters are then selected in the following range listed in Table~\ref{tab:lr1} and Table~\ref{tab:lr2}. For each range, we perform experiment on three values: the lower one, the upper one, and the middle one. Method \textit{MTL} adopt $\nu_{\rm mtl}=0.1.$
\begin{table}[h]
\centering
\caption{Learning rates of experiments.}\label{tab:lr1}
\begin{tabular}{llllll}
\hline
\multicolumn{2}{l}{$K_0$}     &$1$&$3$  & $6$ & $10$ \\ \hline
	DMHP &lr.& $1 \pm .1 \times 10^{-3}$& $3\pm .1 \times 10^{-3}$ & $6.5\pm .1 \times 10^{-3}$ & $7\pm .1 \times 10^{-3}$ \\ \hline
 \multirow{2}{*}{Two Step} &inner lr.& $1 \pm .1 \times 10^{-5}$& $5 \pm .1 \times 10^{-5}$ & $5 \pm .1 \times 10^{-5}$ & $1 \pm .1 \times 10^{-4}$ \\
 &outer lr.& $1 \pm .1 \times 10^{-3}$& $1 \pm .1 \times 10^{-2}$ & $1.5 \pm .1 \times 10^{-2}$ & $1 \pm .1 \times 10^{-2}$ \\ \hline
 HARMLESS &inner lr.& $5 \pm .1 \times 10^{-5}$& $5 \pm .1 \times 10^{-6}$ & $2 \pm .1 \times 10^{-4}$ & $7 \pm .1 \times 10^{-5}$ \\ 
 (MAML) &outer lr.& $6 \pm .1 \times 10^{-4}$& $2 \pm .1 \times 10^{-4}$ & $6 \pm .1 \times 10^{-5}$ & $4.5 \pm .1 \times 10^{-6}$ \\ \hline
 HARMLESS  &inner lr.& $5 \pm .1 \times 10^{-4}$& $1 \pm .1 \times 10^{-5}$ & $3 \pm .1 \times 10^{-5}$ & $1.5 \pm .1 \times 10^{-6}$ \\
 (FOMAML)&outer lr.& $6 \pm .1 \times 10^{-4}$& $2 \pm .1 \times 10^{-4}$ & $6 \pm .1 \times 10^{-5}$ & $4.5 \pm .1 \times 10^{-6}$\\ \hline
\end{tabular}
\end{table}

\begin{table}[h]
\centering
\caption{Learning rates of baseline experiments.}\label{tab:lr2}
\begin{tabular}{ll}
\hline
Method      & Learning Rate \\ \hline
MLE-Sep & $5 \pm .1 \times \times 10^{-5}$  \\
MLE-Com & $1 \pm .1 \times 10^{-3}$            \\ 
MTL &     $1 \pm .1 \times 10^{-3}$       \\ \hline
\end{tabular}
\end{table}


\subsection{Real Datasets}

In this section, we introduce the experimental detail of the real datasets. We run our experiment with same inner and outer learning rate, denoted by $\eta$. For simplicity, we also set $\eta=\eta_{\alpha}=\eta_{\bm{\theta}}$, and search over $\{10^{-4}, 10^{-3}, 10^{-2}, 10^{-1}\} \otimes \{1,2,3,4,5\}$, where the element-wise product of two sets is defined as $A \otimes B = \{ab | a \in A, b \in B\}$.
 We search $K \in \{2, 3, 5\}$ and $\nu_{\rm{mtl}}$ in range $\{0.1,0.01,0.001\}$. We perform grid search over the hyper-parameters, and obtain the candidate models. Then we perform multi-split procedure. 

Because StackOverflow dataset is very large, it is too expensive to perform grid search. To accommodate this, we first split a validation set and a test set, then performing hyper-parameter search by flipping. Each experiment of StackOverflow dataset is run under $5$ different settings.

In Table \ref{tab:real1} we report one of the models that is picked by multi-split procedure. We remark that in most cases, the procedure picks only one model repeatedly.


\begin{table}[h]
	\centering
	\caption{\label{tab:real1}Settings of experiments.}
	\scalebox{0.8}{
	\begin{tabular}{llllll}
		\hline
		data type      &911-Calls  & Linkedin & MathOverflow &StackOverflow \\ \hline
		Baseline 1 &$\eta= 4\times10^{-4}$      & $\eta= 1\times10^{-3}$&   $\eta= 5\times10^{-4}$&   $\eta= 5\times10^{-4}$           \\ 
		Baseline 2& $\eta=  3\times10^{-}  $   & $\eta= 5\times10^{-3}$ & $\eta= 1\times10^{-3}$& $\eta= 1\times10^{-3}$       \\ 
		MTL &$\eta= 3\times10^{-5}, \nu_{\rm{mtl}}=0.1$        &$\eta= 1\times10^{-2}, \nu_{\rm{mtl}}=0.1$  & $\eta= 4\times10^{-4}, \nu_{\rm{mtl}}=0.1$&   $\eta= 5\times10^{-4}, \nu_{\rm{mtl}}=0.1$     \\ 
		DMHP &$\eta= 3\times10^{-5},K=2$& $\eta= 1\times10^{-3},K=3$&$\eta= 4\times10^{-3},K=3$&$N\backslash A$\\
		MAML&$\eta= 3\times10^{-4},K=3$         &  $\eta= 5\times10^{-1},K=3$& $\eta= 3\times10^{-4},K=3$&   $\eta= 1\times10^{-3},K=2$        \\ 
		FOMAML &$\eta= 3\times10^{-5},K=2$        &  $\eta= 1\times10^{-2},K=5$& $\eta= 2\times10^{-4},K=2$&$\eta= 4\times10^{-4},K=3$          \\ 
		Reptile &$\eta= 5\times10^{-3},K=2$        &$\eta= 2\times10^{-1},K=3$ & $\eta= 4\times10^{-2},K=2$&  $\eta= 4\times10^{-2},K=2 $         \\ \hline
	\end{tabular}}
\end{table}

\subsection{Ablation study}

\begin{wraptable}{r}{7.cm}
\centering
\vspace{-10pt}
\caption{Learning rates of experiments of ablation study.}
\begin{tabular}{lr}
\hline
data type                   & LR   \\ \hline
Remove inner heterogeneity ($K=3$)&  $0.1$\\ 
Remove inner heterogeneity ($K=5$)&  $0.1$\\ 
Remove grouping (MAML) & $0.1$ \\
Remove grouping (FOMAML) &  $0.01$\\ 
Remove grouping (Reptile) & $0.2$\\
Remove graph (MAML) &  $0.2$ \\
Remove graph (FOMAML) &  $0.005$\\ 
Remove graph (Reptile) &  $0.2$\\  \hline
\end{tabular}
\end{wraptable}

In this section we introduce the experimental detail of the ablation study. Specifically, the tuning process of the ablation study is as follows: We start from the same setting as the corresponding real experiment in previous section. For example, experiment \textit{Remove graph (FOMAML)} corresponds to HARMLESS (FOMAML). We first use the same learning rate and $K$ as HARMLESS (FOMAML) to perform experiment. If the experiment runs well, we adopt the experiment result. If the training does not converge, we decrease the learning rate and run again.



\end{document}